\documentclass[pdflatex,sn-mathphys-num]{sn-jnl}% Math and Physical Sciences Numbered Reference Style
\usepackage{parskip}        % Better paragraph spacing
\usepackage{amsmath, amssymb}  % Math symbols & environments
\usepackage{graphicx}       % For including images
\usepackage{float}          % Force figure/table positions
\usepackage{caption}        % Better control over captions
\usepackage{subfigure}      % Sub-figures
\usepackage{subcaption}
\usepackage{rotating}       % Rotating figures/tables
\usepackage{pdflscape}      % Landscape pages
\usepackage{lscape}         % Alternative for rotating pages

\usepackage{booktabs}       % Professional tables
\usepackage{longtable}      % Tables spanning multiple pages
\usepackage{multirow}       % Multi-row cells in tables
\usepackage{tabularx}       % Flexible column widths
\usepackage{tabulary}       % Alternative for dynamic-width tables
\usepackage{tabularray}     % Modern tables

\usepackage{algorithm}      % Algorithm floating environment
\usepackage{algpseudocode}  % Pseudocode for algorithms

\usepackage{xcolor}         % For defining custom colors
\usepackage{soul}           % For highlighting text

\usepackage{url}            % Properly format URLs
\usepackage{hyperref}       % Clickable links
\hypersetup{
    colorlinks   = true,
    linkcolor    = blue,
    urlcolor     = blue,
    citecolor    = teal
}

\theoremstyle{thmstyleone}%
\theoremstyle{thmstyletwo}%

\theoremstyle{thmstylethree}%

\begin{document}

\title[Article Title]{An Analytical Framework to Enhance Autonomous Vehicle Perception for Smart Cities}


\author[1,2]{\fnm{Jalal} \sur{Khan}}\email{mjalal@uaeu.ac.ae}
\author[1,2]{\fnm{Manzoor Ahmed} \sur{Khan}}\email{manzoor-khan@uaeu.ac.ae}
\author*[1]{\fnm{Sherzod} \sur{Turaev}}\email{sherzod@uaeu.ac.ae}
\author[1]{\fnm{Sumbal} \sur{Malik}}\email{201990107@uaeu.ac.ae}
\author[1,2]{\fnm{Hesham} \sur{El-Sayed}}\email{helsayed@uaeu.ac.ae}
\author[1]{\fnm{Farman} \sur{Ullah}}\email{farman@uaeu.ac.ae}

\affil[1]{\orgdiv{College of Information Technology}, \orgname{United Arab Emirates University, United Arab Emirates}, \orgaddress{\city{Al Ain}, \country{UAE}}}

\affil[2]{\orgdiv{Emirates Center for Mobility Research}, \orgname{United Arab Emirates University, United Arab Emirates}, \orgaddress{\city{Al Ain}, \country{UAE}}}


\abstract{The driving environment perception has a vital role for autonomous driving and nowadays has been actively explored for its realization. The research community and relevant stakeholders necessitate the development of Deep Learning (DL) models and AI-enabled solutions to enhance autonomous vehicles (AVs) for smart mobility. There is a need to develop a model that accurately perceives multiple objects on the road and predicts the driver's perception to control the car's movements. This article proposes a novel utility-based analytical model that enables perception systems of AVs to understand the driving environment. The article consists of modules: acquiring a custom dataset having distinctive objects, i.e., motorcyclists, rickshaws, etc; a DL-based model (\textit{YOLOv8s}) for object detection; and a module to measure the utility of perception service from the performance values of trained model instances. The perception model is validated based on the object detection task, and its process is benchmarked by state-of-the-art deep learning models' performance metrics from the nuScense dataset. The experimental results show three best-performing YOLOv8s instances based on \textit{mAP@0.5} values, i.e., SGD-based (\textit{0.832}), Adam-based (\textit{0.810}), and AdamW-based (\textit{0.822}). However, the AdamW-based model (i.e., car: \textit{0.921}, motorcyclist: \textit{0.899}, truck: \textit{0.793}, etc.) still outperforms the SGD-based model (i.e., car: \textit{0.915}, motorcyclist: \textit{0.892}, truck: \textit{0.781}, etc.) because it has better class-level performance values, confirmed by the proposed perception model. We validate that the proposed function is capable of finding the right perception for AVs. The results above encourage using the proposed perception model to evaluate the utility of learning models and determine the appropriate perception for AVs.}




\maketitle

\section{Introduction}
\label{sec:introduction}
The intelligent transportation systems (ITS) and autonomous driving (AD) technology have come a long way. With the advancements in machine learning (ML) and deep learning (DL) based technologies, we have new and exciting opportunities to provide ITS services to autonomous vehicles (AVs)~\cite{khan2023augmenting, khan2022level, khan2023advancing, khan2024vehicle}. For instance, a plethora of learning models are benchmarked against the well-known popular and public AD datasets such as KITTI~\cite{geiger2012we, TheKITTI96:online}, nuScenes~\cite{caesar2020nuscenes, nuScenes68:online}, Waymo~\cite{sun2020scalability, OpenData48:online}, and OPENV2V~\cite{xu2022opv2v}, among others. However, this abundance of learning models introduces a new challenge for the perception system of the AV: \textit{the dilemma of perception}. While accurate perceptions in the core perception layer of AD technology could enable more strategic and focused applications of AI-based models in safe and sustainable transport systems, the complex dynamics of urban environments pose challenges in selecting accurate perceptions, which are crucial for achieving a fully autonomous vehicle~\cite{khan2023augmenting, malik2021road}. In addition, the perception system creates a perception service through different tasks, and therefore, even accurate perception will vary in terms of accuracy, performance, and the information needed to execute AV's maneuvers~\cite{khan2022overview}. Therefore, the AVs are faced with the problem of selecting the right perception, which is vital for safe and efficient autonomous driving.

An AV creates the perception of the environment through a perception system, which usually depends on the fundamental tasks of computer vision and refers to the ability to extract relevant data (a.k.a objects) from the surroundings of the driving environment~\cite{malik2023collaborative, malik2023carla}. Based on the perception system results, AVs can make decisions about their future maneuvers~\cite{malik2022autonomous, malik2023should}. The higher the quality of the perception results, the more reliable observations AV can make, enabling accurate decision-making. The existing research works focus on either single-modality or multi-modality solution approaches for perception creation~\cite{khan2022level,mao20233d}. A large number of solution approaches can also be seen as the benchmarked methods for KITTI~\cite{geiger2012we, TheKITTI96:online} and nuScenes~\cite{caesar2020nuscenes, nuScenes68:online} datasets. The relevant research community has been actively proposing perception systems in AD. For instance, SparseFusion~\cite{xie2023sparsefusion}, CMT-NaiveDETR~\cite{yan2023cross}, CBM-Fusion~\cite{song2023spatial}, MSMDFusion-TA~\cite{jiao2023msmdfusion}, Adaptive-Fusion~\cite{liu2022communication}, InterFrame-Fusion~\cite{liu2022brief}, etc., utilize real-world data, and state-of-the-art (SOTA) datasets like nuScenes, KITTI, Waymo~\cite{sun2020scalability}, and others for tasks such as object detection, tracking, planning, and control. In addition, only a few research articles, such as DCNN-Fusion~\cite{zeng2022enabling}, E-DCNN~\cite{lecun2010convolutional}, and WM-YOLO~\cite{kim2019advanced}, have proposed solution approaches to address feature mismatches and missing detection, aiming to achieve accurate perception. In this research, we proposed a novel utility-based perception satisfaction function to allow the perception systems of AVs to select the correct perception. To ensure the AVs are fully satisfied with the given perception service, we need to measure the utility of the performance values of solution approaches that provide accurate perception. Therefore, we formulated an analytical model using a well-known theoretical framework. Our perception satisfaction function considers the perceptual performance values of SOTA learning models and returns utility based on their performance. To imitate real-world use case scenarios, we created numerous instances based on a YOLOv8s DL model.

The DL-based model instances are trained with our custom dataset, and the perception performance values are recorded to validate our perception satisfaction function. Before validation, the learning model instances are divided into four different sets based on their optimizers. A best-performing learning model instance is obtained from each set, and then a single DL-based model instance is selected from these best-performing instances. Our comprehensive process confirms the imitation of the traditional way of accurate perception creation by the learning models available in the literature. Now that we have mimicked the conventional method and obtained an optimal DL-based model instance, we relied on our perception satisfaction function to determine if the optimal learning model instance and the best-performing DL-based model instances actually satisfy the preferences of AVs. Consequently, a bird's-eye-view analysis determined that both optimal model instances and best-performing model instances fully satisfy the perception of AVs. However, a granular analysis revealed that even our optimal fine-tuned (custom-trained over a custom dataset) model instance and the models benchmarked at nuScenes produced poor performance and failed to achieve a delighted state for AVs. With our proposed solution approach, we were able to figure out the reasons behind such poor performances. Therefore, we believe that our proposed satisfaction model can evaluate and measure the satisfaction level of AVs of DL models. Validating DL-based solution approaches from a vehicle satisfaction perspective will enable the research community to develop reliable and responsible ITS services and sustainable transport systems, scaled toward the horizons of AD technology.

The significant contributions of this paper are summarized as follows:
\begin{itemize}
    \item The design of a novel utility-based analytical satisfaction model to enable the AVs to perceive the correct perception of the driving environment.
    \item The acquisition of a new multi-object customized image dataset from urban environments.
    \item The comprehensive DL-based experimental setup to validate the proposed model using the object detection task.
    \item The model tuning approach to select various hyperparameters and find the optimal configuration for a fine-tuned model.
    \item The comparison of the fine-tuned model with SOTA DL models' performance metrics from the nuScense dataset.
    \item The confirmation of AdamW-based instance as the most effective YOLOv8s variant for accurately detecting cars, motorcyclists, and trucks in dynamic environments.
\end{itemize}

The paper is organized into five sections and 1 appendix. Section~\ref{sec:percSatFunc} presents the formulation of the proposed utility satisfaction function, problem domain, system settings, function suitability for realistic scenarios, and perception quantification. Section~\ref{sec:sfValPerTasks} presents the validation of our proposed satisfaction function, dataset acquisition, DL model, and relevant approaches. Section~\ref{sec:expResults} presents the experimental setup and discusses the experiments and results in detail. Section~\ref{sec:concl} presents the conclusion of the research work. The Appendix~\ref{app:details} provides an online link for the class-level performance, the confusion matrices, and Precision-Recall (PR) curves of all YOLOv8 model instances.

\section{Perception Satisfaction Function for AVs}
\label{sec:percSatFunc}
In this section, we focus on modeling the perception satisfaction function. For this, we rely on the well-studied theoretical framework from the field of economy, \textit{utility theory}. Before we start the modeling, in what follows next, we provide a brief primer on the utility theory.

Utility is an abstract concept measuring user satisfaction, originated in the field of economics. It was derived mainly from the works of Von Neumann and Morgenstern~\cite{kuhn1958john}. They designed a function to measure the relative preference of users for different levels of decision metrics. This function is commonly known as the \textit{utility function} and is widely used in economics and decision theory. The concept of a utility function relates to the measurement of user satisfaction. Meaning thereby, the users' relative preferences for different levels of decision metric values can be measured by the function, which is denoted as $U: X \to \mathbb{R}$, represents the preference for all $x$ and $y \in X$ such that $U(x) \ge U(y)$. Hence, the preference relation can be established. For instance, we captured the user satisfaction for service cost in future heterogeneous wireless networks~\cite{khan2011game}. However, it is also worth noting to raise a question once the preference relation is established: \textit{what happens when the satisfaction goes beyond a desired position (threshold) in the perspective of a utility function?} To answer this and understand the effects of additional satisfaction, we need to borrow a relevant concept from economics, i.e., \textit{marginal utility}. Marginal utility follows the principle of diminishing marginal returns, and it measures the additional satisfaction gained from consuming each additional unit of a good or service. The diminishing law refers to the decrease in satisfaction experienced when consuming each successive service after a desired threshold. In the context of AD, marginal utility represents the additional satisfaction or utility gained by the AVs from each unit of the service provided by the core layers, such as perception, planning, and control.

\subsection{Problem Domain and System Settings}
\label{subsec:pdss}
The problem domain is a complex and urban environment for autonomous driving. In this research, we focus mainly on creating and grading perception at the autonomous vehicle level. The core perception layer of AD captures the context of the vehicle's environment~\cite{khan2023augmenting}, which may be carried out through different perception tasks such as detection, tracking, segmentation, and prediction, among others. In addition, the perception can be created through a single or multiple perception tasks. However, the dynamics of complex urban environments pose challenges to achieving accurate perception, which is crucial for a fully autonomous vehicle. Now that the perception is created through different tasks, it will be different in terms of accuracy and the information needed to execute vehicle maneuvers. The vehicle is faced with the problem of selecting the correct perception, which the car will choose based on the proposed service (perception in this case) satisfaction function. We now formally define the problem domain for a finite number of {\it agents (vehicles in our case)} represented by $\mathcal{N}=\{1,2,\ldots,n\},\ n=|\mathcal{N}|.$ Each agent has a finite number of actions (generally arbitrarily large but in the considered settings confined to a few actions), i.e., selection of the right source of perception broadly categorized into that generated by detection, tracking, segmentation, prediction, etc. Now we proceed to model the perception satisfaction function.

\subsection{Proposed Service Satisfaction Function}
\label{subsec:psff}
Let $\mathcal{U}(.)$ be the service satisfaction function of AVs, where the service in this setting is tightly associated with the perception and contextual understanding of the environment. As discussed in Section~\ref{subsec:pdss}, where the problem domain is detailed, such services may be created through different on-board deployed sensors at the vehicle level, i.e., a standalone service offered by a single perception task (denoted by $\mathcal{U}_s(.)$) and a cooperative service offered by multiple perception tasks (denoted by $\mathcal{U}_m(.)$). To avoid confusion in notations while capturing the scope of the satisfaction function, let $\mathcal{U}_i(.)$ be the service satisfaction function for the ego vehicle $i$. It is worth highlighting that $\mathcal{U}_i(.)$ is achieved through an arbitrary relation of perception tasks. The utility function is specific to the context and environmental understanding, which is created through various perception tasks carried out by onboard (on vehicle-deployed) sensors, e.g., Cameras, LiDARs, Radars, etc. Therefore, perception creation is a function of several perception tasks. One of the strengths of our solution approach is enabling the pluggability of off-the-shelf DL-based object detection, tracking, segmentation, and prediction models. Hence, the proposed service satisfaction function for \(\mathcal{V}_i\) is computed using the following Equation~\eqref{psfunc}:

\begin{equation}
\mathcal{U}_i :=
    \sum_{j = 1}^N 
    \omega_j \mathcal{Y}_{jk}(s),  \forall k \in \mathcal{K}
\label{psfunc}
\end{equation}

where $\mathcal{Y}$ captures the quantifiable perception created through different perception tasks carried out by onboard deployed sensors (i.e., represented by the set $\mathcal{K} = \{ \mathcal{D}_i, \mathcal{T}_i, \mathcal{S}_i, \mathcal{P}_i \}$, where $\mathcal{D}_i$ corresponds to the detector (e.g., detection task), $\mathcal{T}_i$ corresponds to the tracker (e.g., tracking task), $\mathcal{S}_i$ corresponds to the segmentor (e.g., segmentation task), $\mathcal{P}_i$ corresponds to the predictor (e.g., prediction task), $s$ corresponds to the type of sensor (e.g., LiDAR, Radar, Camera, etc.). $\omega$ is the associated weight value and the confidence variable. The necessary condition for the confidence value is $\sum_{j=1}^N \omega_j = 1$,  which is dynamically adjusted by the ground truth and AI model validation outcomes. This is to say that the accuracy value will drive the selection of the model in autonomous vehicles. The model selection will positively influence its impact on sustainable, safe, and intelligent transportation systems, as well as smart mobility. It should be noted that for the values to be plugged into the proposed satisfaction function, we rely on the most widely used performance measurement metrics that help quantify perception satisfaction values. Examples of considered metrics include mean average precision (mAP), average multi-object tracking accuracy (AMOTA), mean intersection over union (mIoU), and minimum average displacement error (mADE). The corresponding perception tasks for the considered metrics are represented by the set $\mathcal{K}$. Please refer to Section~\ref{sec:expResults} for more details on performance evaluation and its consequent pluggability into the proposed satisfaction function. Now that we have modeled the perception satisfaction function, we need an appropriate mathematical function for realistic use-case scenarios.

\subsection{Function Suitability for Realistic Scenarios}
\label{subsec:fsfrs}
The mathematical function for perception satisfaction may take different shapes, i.e., linear, step, exponential, concave, and sigmoid. Choosing any of these functions is strictly driven by the parameters under consideration, making its selection a challenge. The utility of satisfaction for vehicle-level perception depends on different perception tasks. In addition, each perception task may produce varying levels of perception. For instance, the perception enabled by a detection task can be achieved by a lower, acceptable, or higher detection performance value (mAP). This is also true for the other performance measurement metrics such as AMOTA, mIOU, mADE, etc. Therefore, we choose a piecewise function, which allows us to have many sub-functions for substituting different perception tasks. Furthermore, the choice of a piecewise function also allows us to design and model different pieces of a particular perception task (object detection). For instance, the performance metric (mAP) values of an object detection task can produce minimum satisfaction, acceptable satisfaction, or maximum satisfaction based on various factors such as threshold value and a particular region in a road segment, among others. 

Based on the minimum and maximum threshold values for specific real-world scenarios, the first component of a particular perception task (function) may result in minimum (zero) satisfaction if it is below or equal to the minimum threshold value. Similarly, the third component of the same perception task (function) may result in maximum satisfaction (five) if it is above or equal to the maximum threshold value. However, the choice of function for the second component may range from linear to exponential to concave, depending on various factors such as safety, efficiency, threshold values, road segment type, complex dynamics, and underlying real-world scenarios. The appropriate mathematical piecewise function for our proposed perception satisfaction function is presented as follows:

\begin{equation}
\mathcal{U}_i = \left. \begin{cases}
0 \quad \quad \quad \quad \quad \quad \quad  \quad \quad \;\;\mbox{if} \quad y^m_{i} \; \leq \underline{y}^m_{i}\\ 
\mu_{i}(y^m_{i})\frac{1 - e^{-\beta_{m}(y^m_{i} - \underline{y}^m_{i})}}{1 - e^{-\beta_{m}(\bar{y}^m_{i} - \underline{y}^m_{i})}} \quad \quad \mbox{if} \quad \underline{y}^m_{i} \; < \;y^m_{i} \; < \bar{y}^m_{i}\\
\mu_{i}(y^m_{i}) \quad \quad \quad \quad \quad \quad  \quad \;\mbox{if} \quad y^m_{i} \;\geq \bar{y}^m_{i}
\end{cases}\right.
\label{pwFunction}
\end{equation}

\begin{figure}[b]
\centering
\includegraphics[width=8cm]{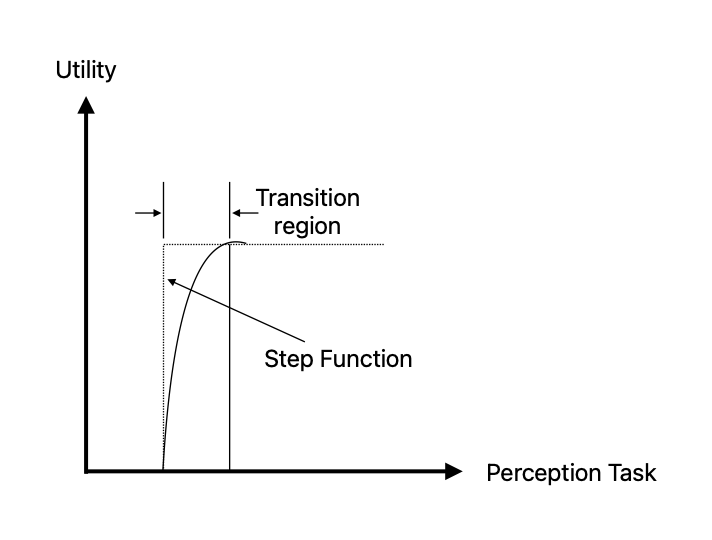}
\caption{The transition region between fully satisfied and fully unsatisfied perception.}
\label{tr}
\end{figure} 

where $y$ is a quantifiable perception created through different perception tasks, $m$ is the performance metrics obtained from the perception tasks, $\mu$ is the threshold value, and $i$ is the ego vehicle performing specific perception tasks. As discussed earlier, the first and third components directly deal with the minimum and maximum threshold values. In contrast, the choice of the second component depends on various behaviors of the function. For instance, it cannot be a linear function due to the complex dynamics of environments for AD. The behavior of a linear function is such that it always produces an increase in the satisfaction utility if the underlying performance metric value increases. On the other hand, the behavior of an exponential function is such that it increases or decreases at a rate proportional to its current performance metric value; in other words, the higher the metric value, the higher the satisfaction utility. Based on a threshold value, the exponential function grows slowly at first and then increases exponentially when the performance metric reaches the threshold. However, let's consider the behavior of a function with concavity, i.e., a concave function. It starts with a slow growth rate, accelerates as the performance metric values increase. Still, the growth rate starts to decelerate after reaching certain threshold values, satisfying the marginal theory (discussed at the start of this section) due to its behavior. Therefore, the appropriate mathematical function for the second component in our Equation~\eqref{pwFunction} is the concave function. However, it is not a good practice to apply concavity to a mathematical function without bounds; therefore, a small transition region is created by the second component as generically shown in Figure~\ref{tr}. The region between two states of perception (fully satisfied, unsatisfied). The length of this region is tuned by $\beta$, such that the higher the value of $\beta$ is, the smaller the region. These range bounds are depicted by the range, say $\underline{y}^m_{i} - \bar{y}^m_{i}$ in Equation~\eqref{pwFunction}.  

\subsection{Perception Quantification}
\label{subsec:pq}
Based on the values of $\mathcal{U}_i$ computed from the above Equations, the service satisfaction for $\mathcal{V}_i$ is quantified into service levels such as \textit{excellent}, \textit{good}, \textit{fair}, \textit{poor}, and \textit{bad} service. These levels of service satisfaction, when expressed numerically, are called \textit{performance score}. The \textit{performance score} is numerical number scaled between $[0-5]$. The average of \textit{performance scores} obtained for the perception task (or multiple tasks) is then known as the \textit{quantified value}. The description of specific numerical value in terms of perception satisfaction is given in Table~\ref{tsv}.

\begin{table}[h] 
\centering
\caption{The Quantified Values}
\label{tsv}
\newcolumntype{C}{>{\centering\arraybackslash}X}
\begin{tabularx}{0.5\columnwidth}{CC}
\toprule
\textbf{Satisfaction Level}	& \textbf{Quantified Value}\\
\midrule
Excellent	& 5	\\
Good		& 4 \\
Fair		& 3 \\
Poor		& 2 \\
Bad 		& 1 \\
\bottomrule
\end{tabularx}
\end{table}

\section{Satisfaction Function Validation for Perception Tasks}
\label{sec:sfValPerTasks}
Our proposed utility function (Equation~\eqref{psfunc}) considers quantifiable perception through various perception tasks performed by onboard sensors at the vehicle level. These perception tasks include, but are not limited to, detection, tracking, segmentation, and prediction of objects in a drivable environment. Each perception task corresponds to a performance metric as discussed in Section~\ref{subsec:psff}. The corresponding perception tasks for the considered performance metrics are represented by the set $\mathcal{K}$. Since the set $\mathcal{K}$ of perception tasks is a collection of domain-specific members, we can rely on the \textit{domain of discourse} that will allow us to use the well-studied quantifier from \textit{mathematical logic} known as \textit{universal quantification}~\cite{Universa49:online}, say $\forall k \in \mathcal{K}$ in Equation~\eqref{psfunc}. In the context of the perception satisfaction function, the universal quantifier expresses that perception can be satisfied by every task (member) of the perception tasks (domain). Therefore, we simplify Equation~\eqref{pwFunction} to solve for targeting a single perception task with multiple instances of the same task and focus on validating the proposed satisfaction function for detection-based perception satisfaction, as shown in Equation~\eqref{odetFunc}.

\begin{equation}
\mathcal{U}_i(d^a_{i}) = \left. \begin{cases}
0 \quad \quad \quad \quad \quad \quad \quad \quad \;\;\;\mbox{if} \quad d^a_{i} \; \leq \underline{d}^a_{i}\\ 
\mu_{i}(d^a_{i})\frac{1 - e^{-\beta_{a}(d^a_{i} - \underline{d}^a_{i})}}{1 - e^{-\beta_{a}(\bar{d}^a_{i} - \underline{d}^a_{i})}} \quad \;\;\mbox{if} \quad \underline{d}^a_{i} \; < \;d^a_{i} \; < \bar{d}^a_{i}\\
\mu_{i}(d^a_{i}) \quad\quad\quad\quad \quad \quad \quad \mbox{if} \quad d^a_{i} \;\geq \bar{d}^a_{i}
\end{cases}\right.
\label{odetFunc}
\end{equation}

Now that we have used \textit{mathematical foundations} to establish our case for object detection-based perception satisfaction, we begin by recording known facts (\textit{data}) about objects in the form of video clips and spatial image structures. Though there exist public datasets for AD, such as KITTI~\cite{geiger2012we}, nuScenes~\cite{caesar2020nuscenes}, Waymo~\cite{sun2020scalability}, OPENV2V~\cite{xu2022opv2v}, etc., these might not address the requirements of a particular object detection task. For instance, rickshaw objects might not be available in the public datasets. Similarly, a well-known Microsoft COCO~\cite{lin2014microsoft} dataset poses a challenge in object detection tasks when a person rides a motorcycle because it allows the detection of everyday objects, such as a person and a bike, independently. Furthermore, achieving novelty and ensuring data quality can only be accomplished with a custom dataset. However, creating a custom dataset is time-consuming and requires expertise in the underlying domain to avoid bias. In what follows, we present our case of data acquisition, dataset creation, data pre-processing, etc.

\subsection{Dataset Acquisition and Data Pre-Processing}
\label{subsec:dataAcq}
\begin{figure*}[]
\centering
\includegraphics[width=12cm]{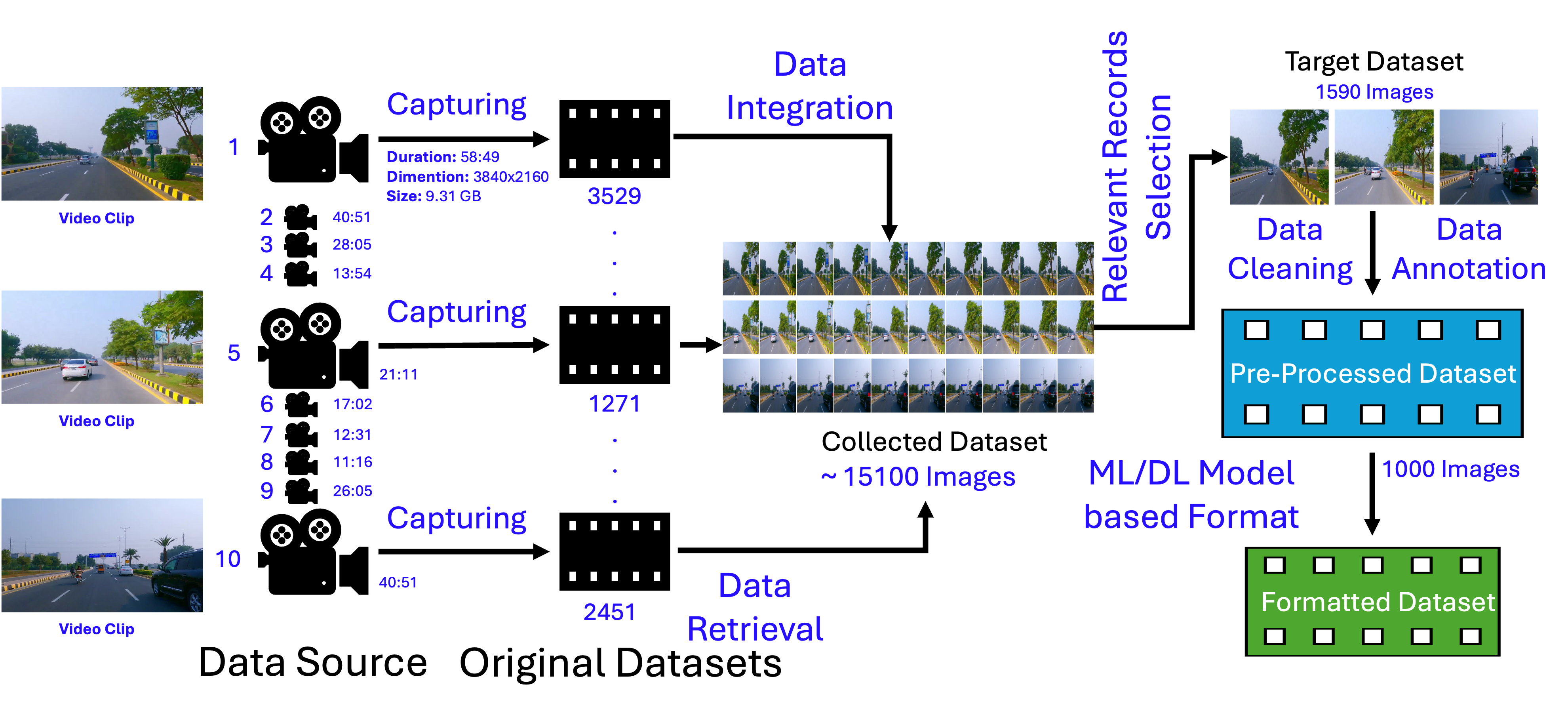}
\caption{Proposed pipeline for data preparation.}
\label{dp}
\end{figure*}

In a typical perception system of AD, the essential aims are to effectively perform perception tasks such as detection, tracking, segmentation, prediction, etc., to allow the autonomous vehicle to plan and control its future maneuvers. For these perception tasks, we need spatial data records. For instance, the detection-based perception requires spatial data to localize and classify the target objects in the drivable environment. Therefore, we capture data of various complex urban road segments in the form of video clips using a high-resolution camera mounted on a vehicle with a rate of 60fps (frames per second) such that the resolution is 3840x2160. A total of 14 video clips were captured with various durations during data collection. A sample of 10 video clips, their duration, and an overall dataset pipeline for data preparation can be seen in Figure~\ref{dp}. Now that we have data sources, we rely on the well-known FFmpeg~\cite{FFmpeg72:online} tool to extract spatial data records (images) from them. We obtained various sizes of original datasets due to differences in the size and duration of the data sources. Next, we integrate the original datasets by getting the necessary data details to form a collected dataset. Consequently, our collected image dataset consists of \textit{15,100} images, where the main objects were cars, persons, motorcycles, tracks, etc.  

Dealing with extensive and unprocessed data records in a collected dataset is cumbersome, and any learning model trained with such a custom dataset will perform poorly. Therefore, we rely on well-known \textit{data pre-processing} from the field of {data mining}. Data pre-processing is used to produce high-quality data outcomes in a speedy and cost-saving manner. There are various pre-processing tasks such as data sampling, variable transformation, handling missing values, dimension reduction, and feature creation, among others. To perform multiple steps of pre-processing on the collected dataset, we use temporal sampling to reduce data volume by focusing on longer-term changes. In addition, we identify regions in the sequence where significant changes occur over a time interval and remove redundant data records, e.g., images of the same objects in the stopping queue of a red signal. In addition, we select relevant and appropriate spatial data records (images) to form our \textit{target dataset}. Consequently, we obtained a spatial target dataset of \textit{1,590} images, which is further filtered through data cleaning, i.e., a data mining process. This process produced a reduction in the dataset size, and a total of \textit{1000} images were selected for the data annotation process, which involves drawing bounding boxes around target objects for perception tasks like object detection. 

Since DL-based models typically perform object detection, they usually label objects with the associated classes based on their training and validation process. Hence, it is imperative to know the main object classes of a dataset. In this connection, we keep a list of main object classes for our target dataset during the data mining processes. The main object classes used for the data annotation process are car, person, motorcycle, truck, and rickshaw. Before the data annotation process, we perform class transformation such that the person class and motorcycle class are transformed into the \textit{motorcyclist} class. It is worth noting that introducing the new motorcyclist class along with the \textit{rickshaw} class brings uniqueness to our custom dataset and presents a challenging environment to the pre-trained DL models. It goes without saying that pedestrians are represented by the person class (to provide a generic view) in large public datasets like COCO~\cite{lin2014microsoft} and nuScenes~\cite{caesar2020nuscenes}. However, this generalization introduces an extra layer of complexity to the DL models, where a human object, whether on the road or a motorcycle, is considered a person class. This opens up unnecessary consequences in the domain of AD. To mitigate this complexity, our proposed motorcyclist class removes the distinction between human objects above the motorcycle and those above the road surface. In this context, we refer to a person without a bike as a pedestrian. Hence, our proposed object classes for data annotation are car, pedestrian, motorcyclist, truck, and rickshaw. 

\begin{table}[h] 
\caption{Objects Count in Custom Dataset}
\label{objCount}
\newcolumntype{C}{>{\centering\arraybackslash}X}
\begin{tabularx}{\columnwidth}{lCCCC}
\toprule
\textbf{Classes}	& \textbf{Train Set}& \textbf{Valid Set}& \textbf{Test Set}& \textbf{Objects}\\
\midrule
car	            & 2466	& 660	& 367	& 3493	\\
motorcyclist	& 977   & 293	& 130	& 1400	\\
pedestrian		& 701   & 197	& 88	& 986	\\
rickshaw		& 252   & 89	& 39	& 380	\\
truck 		    & 773   & 225	& 111	& 1109	\\
\textbf{Objects}  & 5169  & 1464  & 735   & \textbf{7368}  \\
\bottomrule
\end{tabularx}
\end{table}

During the annotation process, manual labeling was performed on the target dataset. We generate the ground truths of spatial data records by storing the bounding boxes around objects. After the annotation process, our pre-processed dataset of \textit{1,000} spatial data records was ready. A total number of \textit{7,368} object instances were annotated in our custom dataset, as shown in Table~\ref{objCount}. The pre-processed dataset is then used to convert the annotation format for the DL models considered. For instance, the YOLO~\cite{ultralyt51:online} based DL models require pre-processed datasets with the YOLO format of annotations. Therefore, we format data into an acceptable form to prepare a formatted dataset for performing object detection tasks of the perception system. Moreover, the formatted data is required in three different sub-datasets, i.e., the training dataset, the validation dataset, and the testing dataset. In this connection, we rely on a well-known split ratio of 70:20:10, which means the training dataset occupies 70\% of the data records, the validation dataset occupies 20\% of the data records, and the testing dataset occupies 10\% of the data records. Now that we have our spatial data records ready for the detection-based perception task, we need an SOTA DL model to be trained over our custom dataset.   

\subsection{The YOLO (You Look Only Once) Model}
\label{subsec:dlm}
Deep learning models are well-known for using deep neural network (layer-based) architectures to learn hidden representations from data to make decisions (predictions). For instance, the object detector's decision to localize and classify an object as either a pedestrian or a vehicle is based on the deep-layered learning of the model. In AD, we often want to have DL-based models for detection-based perception systems. This way, we will allow AVs to drive and navigate without human intervention. However, the solution approaches by DL-based object detection models have expanded so extensively that it is impractical to explore every model~\cite{mao20233d}. Even though if we apply simple inclusion criteria to select and filter promising object detectors, we find that most of them do not work with real-time data, do not generalize over new data records, have no family history, lack in the number of model versions, focus on either speed or efficiency (in terms of accuracy), etc. The filtration process led us to focus on the most updated and current SOTA real-time object detection model, You Only Look Once (YOLOv8)~\cite{ultralyt51:online}, for a perception system in AD. The YOLOv8 has a family history (i.e., YOLOv1, YOLOv2, YOLOv3, YOLOv4, YOLOv5, YOLOv6, and YOLOv7), and it has five different versions (YOLOv8n, YOLOv8s, YOLOv8m, YOLOv8l, YOLOv8x), and it was recently released in January 10th, 2023. Moreover, it surpasses all its previous family members in terms of speed and detection accuracy. Therefore, it has been used to detect common traffic objects~\cite{lin2023traffic} in mixed traffic environments~\cite{afdhal2023real}. In what follows, we leverage the capabilities of the YOLOv8 model for object detection using our custom dataset. 

\begin{figure*}[]
\centering
\includegraphics[width=10.5 cm]{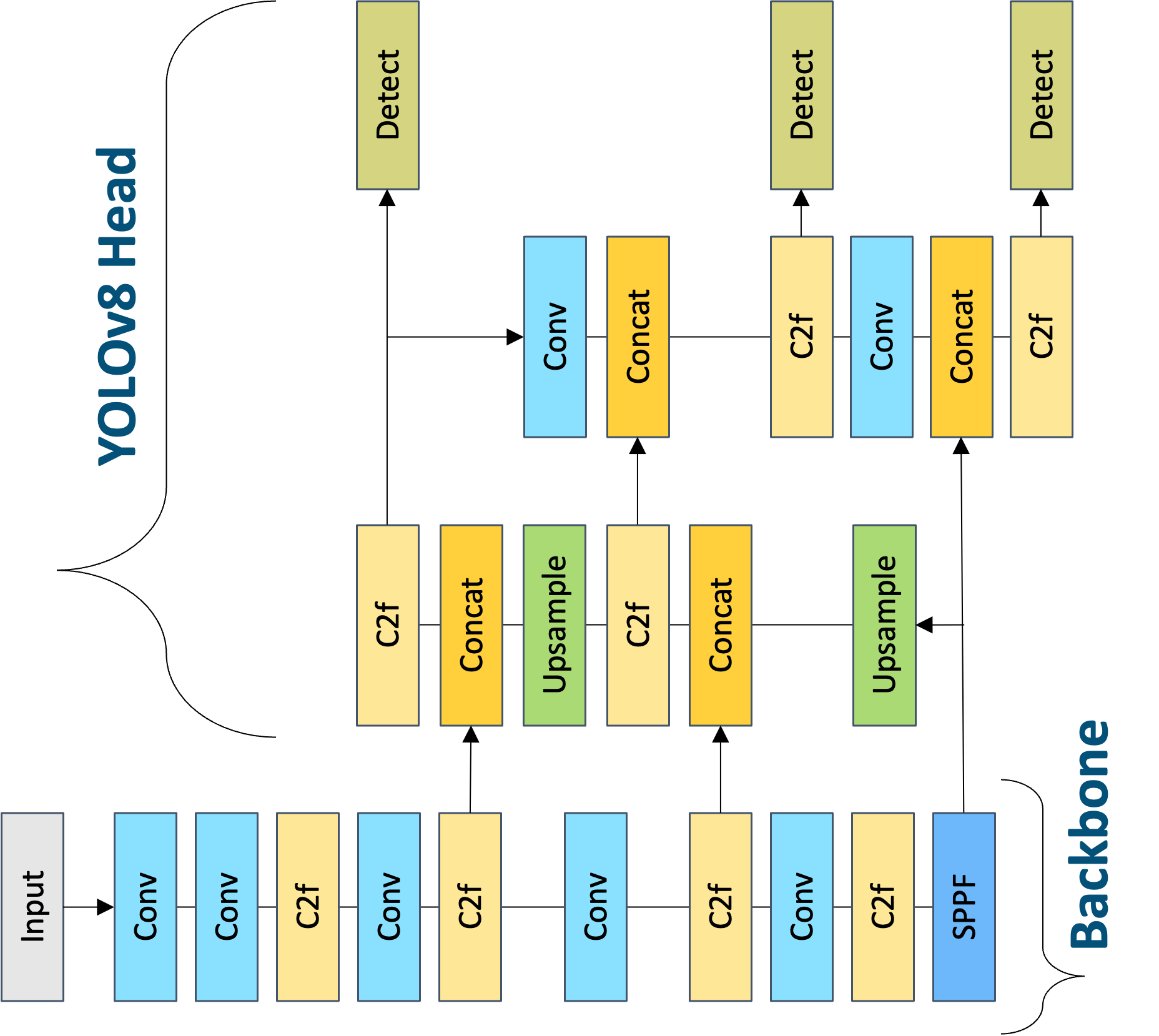}
\caption{The YOLOv8 Architecture.}
\label{y8archi}
\end{figure*}

The YOLOv8 architecture includes a Backbone, Head, SPPF layer, Upsample layers, convolutional layers, linear layers, C2f modules, Detection module, etc. A simplified version of the architecture is shown in Figure~\ref{y8archi}. The \textit{Backbone} ingests a spatial data record (image) from the formatted pre-processed dataset and extracts relevant features by using several convolutional layers. The features are processed through the SPPF layer at different depths, widths, and ratios according to the underlying version of the YOLOv8 model. Therefore, the Backbone of the architecture can generate feature maps at various scales. Now that the feature maps are ready, they are provided to the \textit{Head} of the architecture. The Head takes these feature maps and gives them to the Upsample layers inside the Head to increase their resolution. Moreover, the Head allows each branch to perform its respective task independently from other branches. The design and behavior of the Head help the model to improve its speed and object detection accuracy by maximizing the overall performance of YOLOv8. Therefore, the number of channels and kernel size of each layer are carefully selected. In addition, the C2f modules improve object detection accuracy by combining high-level features with contextual information. Similarly, the Smooth L1 loss function is used for the box regression loss, the sigmoid function is used as the activation function, and the softmax function is used for class probabilities. The detection module maps the high-dimensional features to produce the final output in the form of bounding boxes and object class probabilities. It is noteworthy that the YOLOv8 model directly detects objects and uses an anchor-free approach similar to the YOLOX model~\cite{ge2021yolox}. These details are just a tip of the YOLOv8 iceberg, and therefore, we encourage the reader of this paper to visit~\cite{Briefsum66:online} for a complete model structure and relevant details.  

To further optimize the performance of our considered SOTA real-time object detection model, we rely on a well-known \textit{hyperparameter tuning} approach. It is commonly used to systematically find a combination of hyperparameters, such as epochs, batch size, optimizer, learning rate, stopping criteria, etc., to produce high object detection performance. However, we aim to adopt the hyperparameter tuning approach for two primary reasons: (i) it can fine-tune our considered real-time DL-based object detection model; (ii) it will be instrumental in creating a large number of instances of the underlying DL model, as changing or updating each specific hyperparameter will result in a different instance of the learning model. This is important since we need to imitate real-world scenarios and require a large number of DL models to validate our perception satisfaction function. Therefore, we rely on the distinct instances (versions) of the YOLOv8s model to incorporate their performance metric values, obtained during the hyperparameter approach, into our mathematical piecewise perception satisfaction function. 

To achieve this aim, we consider YOLOv8s with three important hyperparameters with various values: optimizer (SGD, Adam, RMSProp, AdamW); batch-size (16, 32); and epochs (100, 150, 200). The rest of the hyperparameters are the default parameters of YOLOv8. The set of optimizers includes efficient algorithms used to train the underlying DL model instances. The Stochastic Gradient Descent (SGD) belongs to the family of gradient descent algorithms. However, SGD estimates the gradient of the cost function using a few randomly chosen data points. SGD can train learning models that can be generalized. However, they are slow and require regularization to avoid overfitting. On the other hand, Adam (Adaptive Moment Estimation) is a combination of AdaGrad (Adaptive Gradient Algorithm) and RMSProp (Root Mean Square Propagation). While it has a fast convergence speed as compared to SGD, it suffers from a lack of generalization capabilities. Hence, the trained model may not be able to generalize to the unseen data. The RMSProp is included in the optimizer set because we were interested in knowing if RMSProp (standalone) performs well when competing with SGD and Adam. Lastly, we select Adam with weight (AdamW) decay regularization because its convergence speed matches that of Adam, and it enhances the model's generalization capabilities. With these four optimizers, two batch sizes, and three epoch ranges, we prepared a set of \textit{24} different instances of the YOLOv8s model. The performance metrics (mAP@0.5) of these instances are compared to filter out the fine-tuned hyperparameter configuration and are plugged into the detection-based perception satisfaction function. Now that we have all the cases available, we start with the training and testing of the real-world object detection model. 

\subsection{Training and Testing Approaches}
\begin{figure*}[]
\centering
\includegraphics[width=10.5 cm]{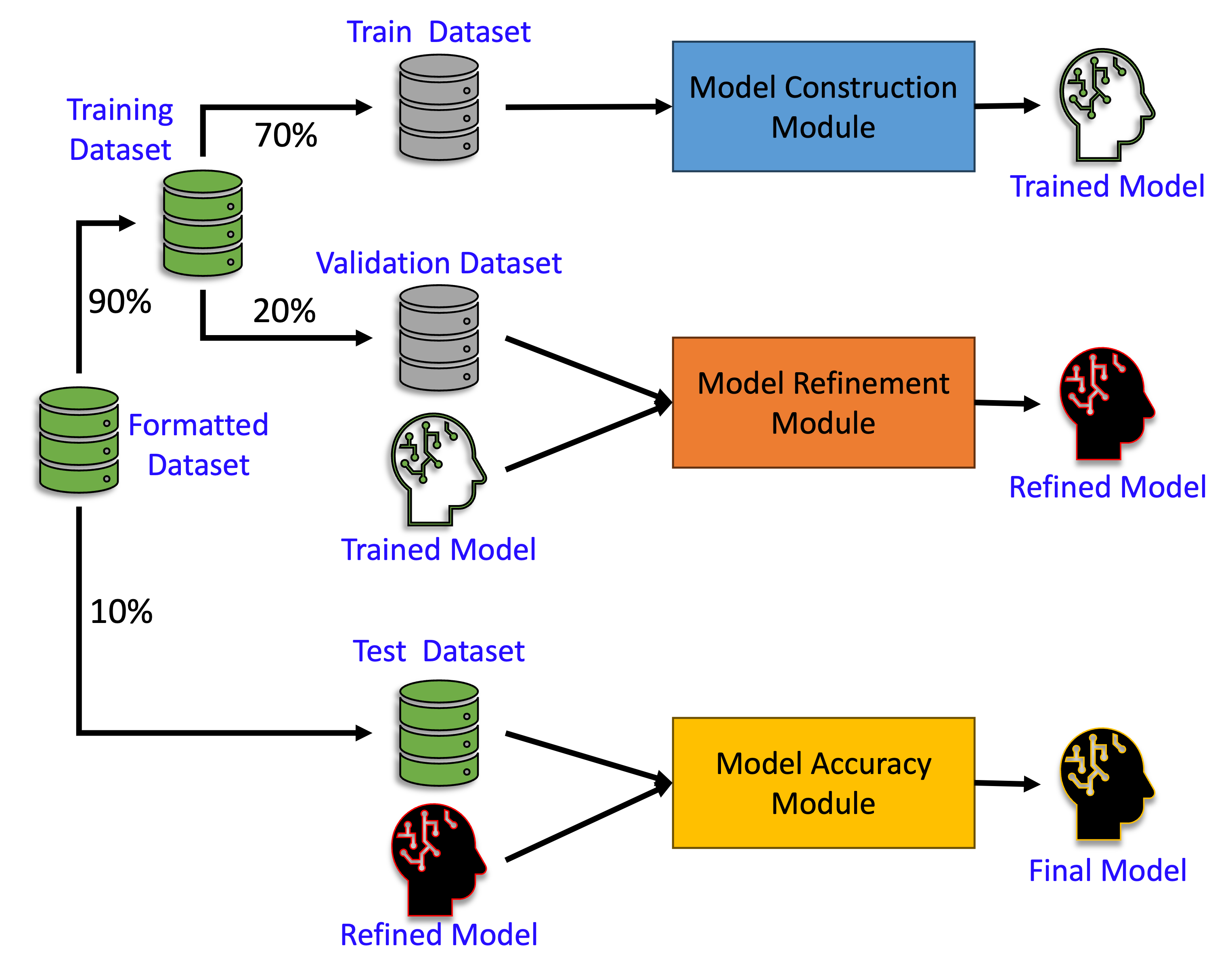}
\caption{Proposed pipeline for training, validation, and testing datasets.}
\label{ptvt}
\end{figure*}

The proposed pipeline for training, validation, and testing can be seen in Figure~\ref{ptvt}. The formatted and pre-processed annotated dataset is split into training and test datasets with a 90:10 ratio, correspondingly. The 90\% training dataset is further divided into train and validation datasets with a 70:20 ratio, respectively. At this stage, we start training each instance of the YOLOv8s model from the 24 instances' set with the custom train dataset. The model construction module in Figure~\ref{ptvt} ingests the training dataset and each instance of the YOLOv8s model with specific hyperparameter configuration and produces a trained YOLO-based DL model. This process is executed for the whole set of 24 different instances of the YOLOv8s model. Consequently, we trained 24 different YOLOv8 model instances. Next, we start with refining the trained model instances. The model refinement module ingests the validation dataset, which was not known to the trained model, and each instance of the trained YOLOv8s model. When we finished refining all models, we had 24 different refined model instances of YOLOv8s. Lastly, we start with finalizing the refined instances of the YOLOv8s model. For this, we use the model accuracy module and ingest the test dataset along with each instance of the refined YOLOv8s model. Consequently, we obtained the inference results for 24 final instances of the YOLOv8s model. This allows us to use a set of 24 different instances of the YOLOv8s-based DL model in our proposed function. Now that we have all the ingredients for our utility function validation, we show the experimental work and results of the research.

\section{Experiments and Results}
\label{sec:expResults}
We performed experiments over a set of YOLOv8s-based DL model instances. The instances of YOLOv8s consider various hyperparameters, i.e., optimizers, batch sizes, and epochs, to obtain performance metrics for object detection-based perception systems in AVs. In what follows, we provide details for our experimental setup, evaluation metrics, and results.

\subsection{Experimental Setup}
\label{subsec:expSetup}
The experiments were conducted using the NVIDIA \textit{Tesla V100 GPU} model, 16.0GB of GPU RAM, 12.7GB of System RAM, and 166.8GB size of Disk space through \textit{Google Colab Pro} platform. The NVIDIA Tesla V100 GPU was accessed through Google Chrome (version 117.0) installed on a physical machine with an Intel Quad-Core i7@ 3.4GHz processor, 32GB of RAM, NVIDIA GeForce GTX 680MX Graphics, and a 27-inch (2560x1440) built-in Display. The software stack includes Python v.3.11.4 (the newest major release) and its family of libraries, such as PyTorch, glob, etc. The YOLOv8s-based DL model instances were implemented through \textit{Ultralytics} package~\cite{ultralyt88:online} using the latest version of the Python programming language. The learning model instances were trained on an image size of 640x640, and different configurations were applied during the hyperparameter fine-tuning stage.

\subsection{Evaluation Metrics}
\label{subsec:evalMetrics}
In this section, we focus on evaluation standards for measuring the performance of our DL-based models. For this, we rely on the well-studied fundamental components of \textit{confusion matrix} from the field of data mining and machine learning. These components are: \textit{TP} (True Positive), \textit{FP} (False Positive), \textit{TN} (True Negative), and \textit{FN} (False Negative). They are used to measure the performance of a learning model, i.e., how accurately a model distinguishes between the classes. The different combinations and relationships of these components result in other essential evaluation metrics, i.e., \textit{precision}, \textit{recall}, \textit{F1-score}, \textit{intersection-over-union} (IoU), \textit{mean average precision} (mAP), etc., which lead us to another level of understanding the holistic view of our learning model's reliability and generalization capability.

\textbf{Precision.} Precision answers the following basic question: How many of the positive predictions were \textit{actually correct} from all positive predictions made by the learning model? To answer this, we rely on precision in Equation~\eqref{precision}: 

\begin{equation}
 \text{Precision} = \frac{\text{TP}}{\text{TP + FP}}
\label{precision}
\end{equation}

\textbf{Recall.} Recall answers the following basic question: How many actual positive instances were \textit{correctly predicted} by the model from all actual positive instances in the data? To answer this, we rely on recall in Equation~\eqref{recall}: 

\begin{equation}
 \text{Recall} = \frac{\text{TP}}{\text{TP + FN}}
\label{recall}
\end{equation}

\textbf{F1-score.} F1-score answers the following basic question: How well does the learning model \textit{balance} between its precision and recall? To answer this, we rely on the F1-score in Equation~\eqref{f1score}: 

\begin{equation}
 F1 = 2 \times \frac{\text{Precision} \times \text{Recall}}{\text{Precision} + \text{Recall}}
\label{f1score}
\end{equation}

\textbf{Mean Average Precision (mAP).} mAP answers the following basic question: How consistently accurate are the learning model's predictions across different recall levels and across all classes? To answer this, we rely on mAP in Equation~\eqref{mAPequ}: 

\begin{equation}
mAP = 
    \frac{1}{N} 
    \sum_{i = 1}^N 
    AP_i
\label{mAPequ}
\end{equation}

where $N$ represents the set of classes, i.e., car, motorcyclist, pedestrian, rickshaw, and truck, and $AP_i$ represents the average precision of class $i$. It is worth noting that the AP for each class relies on \textit{IoU criterion} and its thresholds (a standard threshold in object detection is 0.5 (50\%)). Hence, \textit{mAP@0.5} evaluates the learning models' performance at 50\% of detected bounding box overlap to the ground truth bounding box. Now that we have the most essential evaluation standards, we present our experimental results and provide a discussion to equip the readers with a comprehensive understanding.

\begin{table}[b]
\caption{Comparison of all \textit{mAP@0.5} achieved against Epochs and Batch-Sizes.}
\label{comparisonAllmAPs}
\newcolumntype{C}{>{\centering\arraybackslash}X}
\begin{tabularx}{\columnwidth}{lCCCCC}
\toprule
\multirow{2}{*}{\textbf{Epochs}} & \multirow{2}{*}{\textbf{Batch Sizes}} & \multicolumn{4}{c}{\textbf{Optimizers}} \\
& & \textbf{SGD} & \textbf{Adam} & \textbf{RMSProp} & \textbf{AdamW} \\
\midrule
\multirow{2}{*}{100} & 16 & 0.815 & 0.804 & 0.00013 & 0.804 \\
& 32 & 0.826 & 0.793 & \textbf{0.242} & 0.804 \\
\multirow{2}{*}{150} & 16 & \textbf{0.832} & \textbf{0.810} & 0.0001 & 0.810 \\
& 32 & 0.825 & 0.791 & 0.000006 & 0.810 \\
\multirow{2}{*}{200} & 16 & 0.820 & 0.804 & 0.000004 & 0.810 \\
& 32 & 0.819 & 0.802 & 0.0485 & \textbf{0.822} \\
\bottomrule
\end{tabularx}
\end{table}

\subsection{Experimental Results and Discussions}
\label{subsec:expResultsAndDiscussions}
The YOLOv8-based DL model instances were trained on our custom-formatted dataset. The performance of model instances depends on the quality of data records, which can be improved by thoroughly labeling the objects and accurate class representations in the dataset~\cite{otgonbold2022shel5k}. In our custom dataset, we combined two object classes from the COCO~\cite{lin2014microsoft} dataset, i.e., the \textit{person} class and the \textit{motorcycle} class, into the \textit{motorcyclist} class and introduced a new \textit{rickshaw} class. Therefore, our custom-formatted dataset has a set of five classes, i.e., car, motorcyclist, pedestrian, rickshaw, and truck. The performance (in terms of \textit{mAP@0.5}) of all YOLOv8s-based instances on the proposed custom dataset is summarized in Table~\ref{comparisonAllmAPs}. It shows four different sets (based on the optimizer) of performance results, where each set (from SGD to AdamW) considers different epochs and batch sizes during their training stage. From a bird's-eye view analysis, the trained YOLOv8s model instances showed promising performance for object detection, i.e., the SGD set achieved mAP@0.5 values ranging from 0.815 to 0.832, the Adam set achieved mAP@0.5 values ranging from 0.791 to 0.810, and the AdamW set achieved mAP@0.5 values ranging from 0.804 to 0.822. However, the learning model instances in RMSProp set achieved mAP@0.5 values ranging from 0.000004 to 0.242. The main reason behind the poor performance of the RMSProp-based YOLOv8s model instance is dynamic learning rate adjustments for each weight during training~\cite{RMSpropE67:online}. Therefore, RMSProp is used as an optimization technique in combination with other methods, e.g., the Adam technique. Moreover, its inclusion in the hyperparameter tuning process highlights the importance of model fine-tuning and why we should not choose every learning model and its parameters.

\begin{table*}[b] 
\caption{Comparison of the YOLOv8 instances for Object Detection.}
\label{comparisonAllInstances}
\newcolumntype{C}{>{\centering\arraybackslash}X}
\begin{tabularx}{\textwidth}{lCCCCC}
\toprule
\textbf{Instances}	& \textbf{Precision} & \textbf{Recall} & \textbf{F1} & \textbf{mAP@0.5} & \textbf{mAP@0.5:0.95} \\
\midrule
YOLOv8s-1	&0.845	&0.724	&0.780	&0.815	&0.605 \\
YOLOv8s-2	&0.840	&0.710	&0.760	&0.826	&0.611 \\
\textbf{YOLOv8s-3}	&\textbf{0.855}	&\textbf{0.743}	&\textbf{0.790}	&\textbf{0.832}	&\textbf{0.618} \\
YOLOv8s-4	&0.838	&0.729	&0.780	&0.825	&0.606 \\
YOLOv8s-5	&0.795	&0.766	&0.780	&0.820	&0.611 \\
YOLOv8s-6	&0.843	&0.729	&0.780	&0.819	&0.606 \\
YOLOv8s-7	&0.829	&0.703	&0.760	&0.804	&0.577 \\
YOLOv8s-8	&0.803	&0.711	&0.750	&0.793	&0.568 \\
\textbf{YOLOv8s-9}	&\textbf{0.783}	&\textbf{0.738}	&\textbf{0.750}	&\textbf{0.810}	&\textbf{0.580} \\
YOLOv8s-10	&0.798	&0.698	&0.740	&0.791	&0.565 \\
YOLOv8s-11	&0.795	&0.732	&0.760	&0.804	&0.574 \\
YOLOv8s-12	&0.790	&0.731	&0.750	&0.802	&0.573 \\
YOLOv8s-13	&0.00027	&0.00152	&0.00046	&0.00013	&0.00004 \\
\textbf{YOLOv8s-14}	&\textbf{0.351}	&\textbf{0.201}	&\textbf{0.240}	&\textbf{0.242}	&\textbf{0.115} \\
YOLOv8s-15	&0.00015	&0.00394	&0.00028	&0.0001	&0.00003 \\
YOLOv8s-16	&0.00001	&0.00121	&0.00002	&0.000006	&0.000001 \\
YOLOv8s-17	&0.000008	&0.00137	&0.00001	&0.000004	&0.0000004 \\
YOLOv8s-18	&0.0703	&0.0418	&0.0524	&0.0485	&0.023 \\
YOLOv8s-19	&0.812	&0.719	&0.760	&0.804	&0.58 \\
YOLOv8s-20	&0.832	&0.702	&0.760	&0.804	&0.58 \\
YOLOv8s-21	&0.808	&0.715	&0.760	&0.810	&0.589 \\
YOLOv8s-22	&0.827	&0.715	&0.760	&0.810	&0.591 \\
YOLOv8s-23	&0.775	&0.756	&0.760	&0.810	&0.591 \\
\textbf{YOLOv8s-24}	&\textbf{0.842}	&\textbf{0.728}	&\textbf{0.780}	&\textbf{0.822}	&\textbf{0.594} \\
\bottomrule
\end{tabularx}
\end{table*}

The mAP@0.5 performance values provide a high-level analysis. In Table~\ref{comparisonAllInstances}, another level of detailed analysis can be seen by the precision, recall, F1-score, and mAP@0.5:0.95 (with a step of 0.05) against the given mAP@0.5 performance values. It is important to map these results with the Table~\ref{comparisonAllmAPs}, therefore, the first six instances (YOLOv8s-1 to YOLOv8s-6) form the SGD set, the following six cases (YOLOv8s-7 to YOLOv8s-12) form Adam set, the following six instance (YOLOv8s-13 to YOLOv8s-18) form RMSProp set, and the last six instances (YOLOv8s-19 to YOLOv8s-24) form AdamW set. 

\subsubsection{The YOLOv8s-1 Model Instance}
\label{subsubsec:yolov8_1_model}
The YOLOv8s-1 instance (first tuple in Table~\ref{comparisonAllInstances}) has achieved a mAP@0.5 of 0.815 with precision of 0.845, recall of 0.724, F1-score of 0.780, and mAP@0.5:0.95 of 0.605. These performance values answer the actually correct predictions (precision), correctly predicted actual positive instances (recall), balance between the first two performances (F1), and performance required for critical use-case scenarios (mAP@0.5:0.95). It is worth noting that the mAP@0.5:0.95 value 0.605 drops the learning model instance performance from \textit{81.5\%} to \textit{60.5\%}. Meaning thereby, our learning model is good in making approximate predictions, but not as good when we need more precise predictions required for critical use case scenarios. The main reason for the model's behavior is the working principle of mAP@0.5:0.95, which averages the mAP values calculated at multiple IoU thresholds (from 0.5 to 0.95 with an increment of 0.05). Having said that, a holistic understanding of the DL model instances is achieved by delving into a deeper level of their performance. To delve deeper into analyzing the performance values, we rely on granular analysis and examine the individual class performance of the \textit{YOLOv8s-1} instance, as provided in Table~A1, Appendix~\ref{app:details}.

Consider the first tuple under \textit{YOLOv8s-1: SGD-E100-B16} in Table~A1, the precision value 0.859 indicates that \textit{this} (YOLOv8s-1) DL-based model instance can detect a car as significant as \textit{85.9\%}. The recall value 0.851 indicates that the learning model instance can detect a substantial proportion (\textit{85.1\%}) of the actual cars present in our custom dataset. Similarly, the mAP@0.5 value 0.916 indicates that the learning model instance has high performance (\textit{91.6\%}) at an \textit{IoU} threshold of 0.5. However, the mAP@0.5:0.95 value 0.713 drops the learning model instance performance from \textit{91.6\%} to \textit{71.3\%}. The similar reasons mentioned in the above paragraph apply to this behavior of the model instance. Now that we have a deeper understanding of the performance values, we consider precision, recall, mAP@0.5, and mAP@0.5:0.95 of the YOLOv8s-1 model instance. This allowed us to notice that the optimal precision value (0.921) is for the object motorcyclist, the recall value (0.851) is for the car, the mAP@0.5 value (0.916) is for the vehicle, and the mAP@0.5:0.95 value (0.713) is for the car. Therefore, we propose that the YOLOv8s-1 model instance is a suitable learning model for detecting \textit{car} objects and also \textit{motorcyclist} objects.

To support our proposition, we rely on a well-studied \textit{confusion matrix} from the field of machine learning. The performance values reported in the confusion matrix, as shown in Figure~A1a in Appendix~\ref{app:details}, confirm that the YOLOv8s-1 model instance detected \textit{car} as car for 88\% of its proportion in the custom dataset, whereas the remaining 12\% of its proportion is divided into \textit{truck} and \textit{no detection}. Similarly, the second highest detection (83\%) is performed for \textit{motorcyclist} objects, whereas the remaining 17\% is divided between \textit{car}, \textit{pedestrian}, and \textit{no detection}. A reflection of these results can also be witnessed in the Precision-Recall (PR) curves shown in Figure~A1c in Appendix~\ref{app:details}. It is worth noting that the poor performance of the considered learning model instance for \textit{pedestrian} and \textit{rickshaw} classes is not because of wrong class detections, but because the YOLOv8s-1 model instance was unable to detect them, as indicated by 0.30 (for pedestrian class) and 0.26 (for rickshaw class). Since we have performed a comprehensive \textit{fine-tuning} of the YOLOv8s model, we found that if we change the \textit{optimizer} from SGD to AdamW, and the number of \textit{epochs} is increased from 100 to 200, the detection will increase from 0.69 to 0.75 (75\% of rickshaw objects are detected). The no-detection value will decrease from 0.26 to 0.16, indicating improvements in model performance. This improved version (when focusing on objects of rickshaw class for a particular use-case scenario) of the model instance can be seen as YOLOv8s-23 in Table~\ref{comparisonAllInstances} and as YOLOv8s-23: AdamW-E200-B-16 in Table~A3 in Appendix~\ref{app:details}. Its confusion matrix and PR-curves can be seen in Figure~A5c and Figure~A5f, respectively. At this stage, we conclude the comprehensive analysis of a YOLOv8s-1 DL model instance from its mAP@0.5 value to its components of the confusion matrix and encourage the readers to map any other YOLOv8s-x model instance similarly based on their required or considered use-case scenarios. For this, the readers can rely on the class performance of all YOLOv8s instances in Table~A1, Table~A2, Table~A3 in Appendix~\ref{app:details}. Their confusion matrix and PR-Curves shown in Figure~A1, Figure~A2, Figure~A3, Figure~A4, Figure~A5, Figure~A6 in Appendix~\ref{app:details}. In what follows, we will provide a high-level analysis of the overall comprehensive set of experiments.

\begin{figure*}[b]
    \centering
    \subfigure[Confusion Matrix.\label{fig:cm_sgd_e150_b16}]{\includegraphics[width=0.50\textwidth]{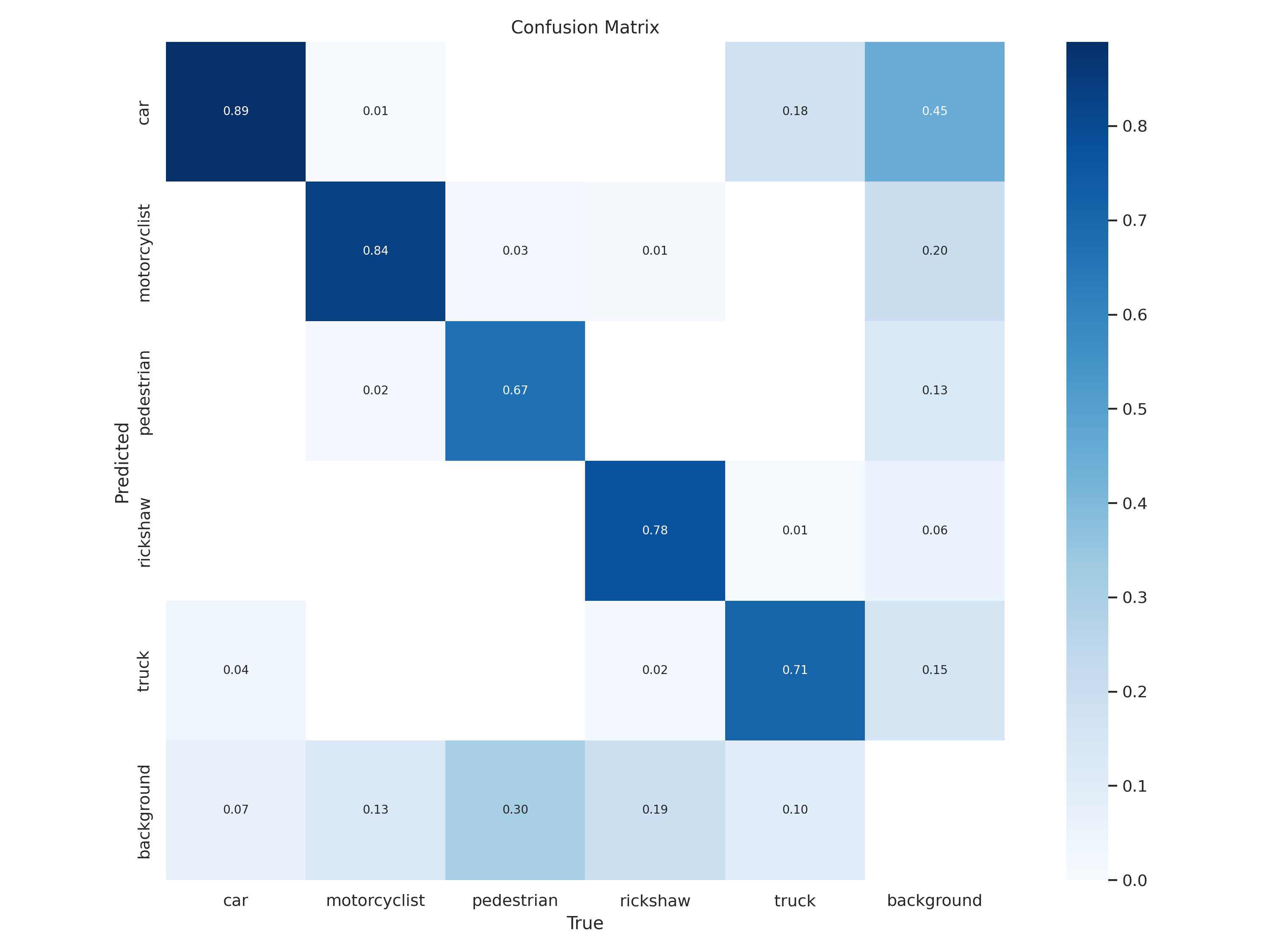}} \hfill
    \subfigure[Precision-Recall Curve.\label{fig:sdg_e150_b16}]{\includegraphics[width=0.40\textwidth]{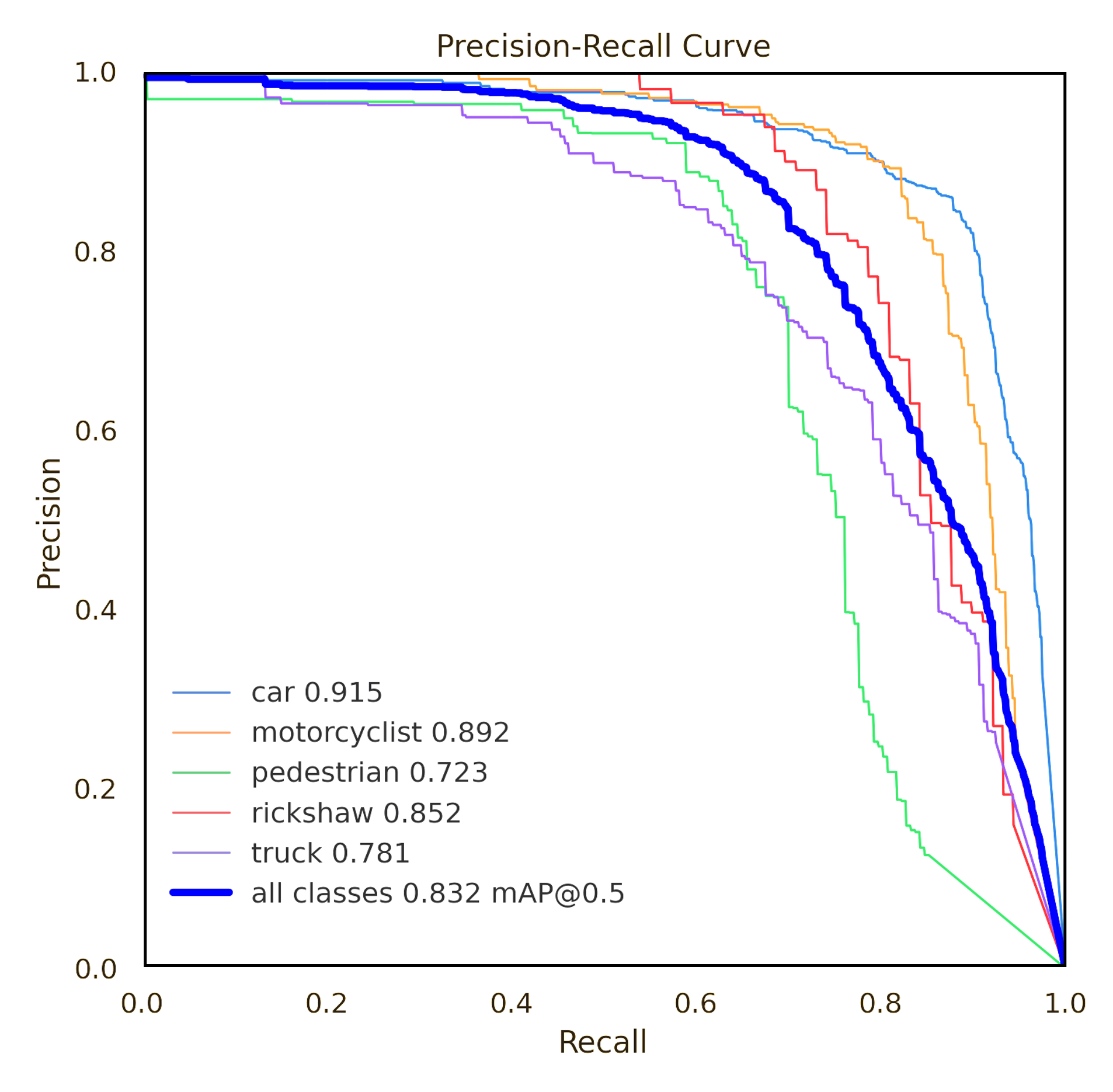}} \\[1ex]

    \caption{The Optimal YOLOv8s instance from SGD Group with 150 Epochs and 16 Batch-size.}
    \label{fig:sgd_150_16}
\end{figure*}

\subsubsection{The Fine-tuned YOLOv8s Model Instance}
\label{subsubsec:fine_tuned_yolov8_model}
\begin{figure*}[t]
    \centering
    \subfigure[Confusion Matrix.\label{fig:cm_adam_e150_b16}]{\includegraphics[width=0.50\textwidth]{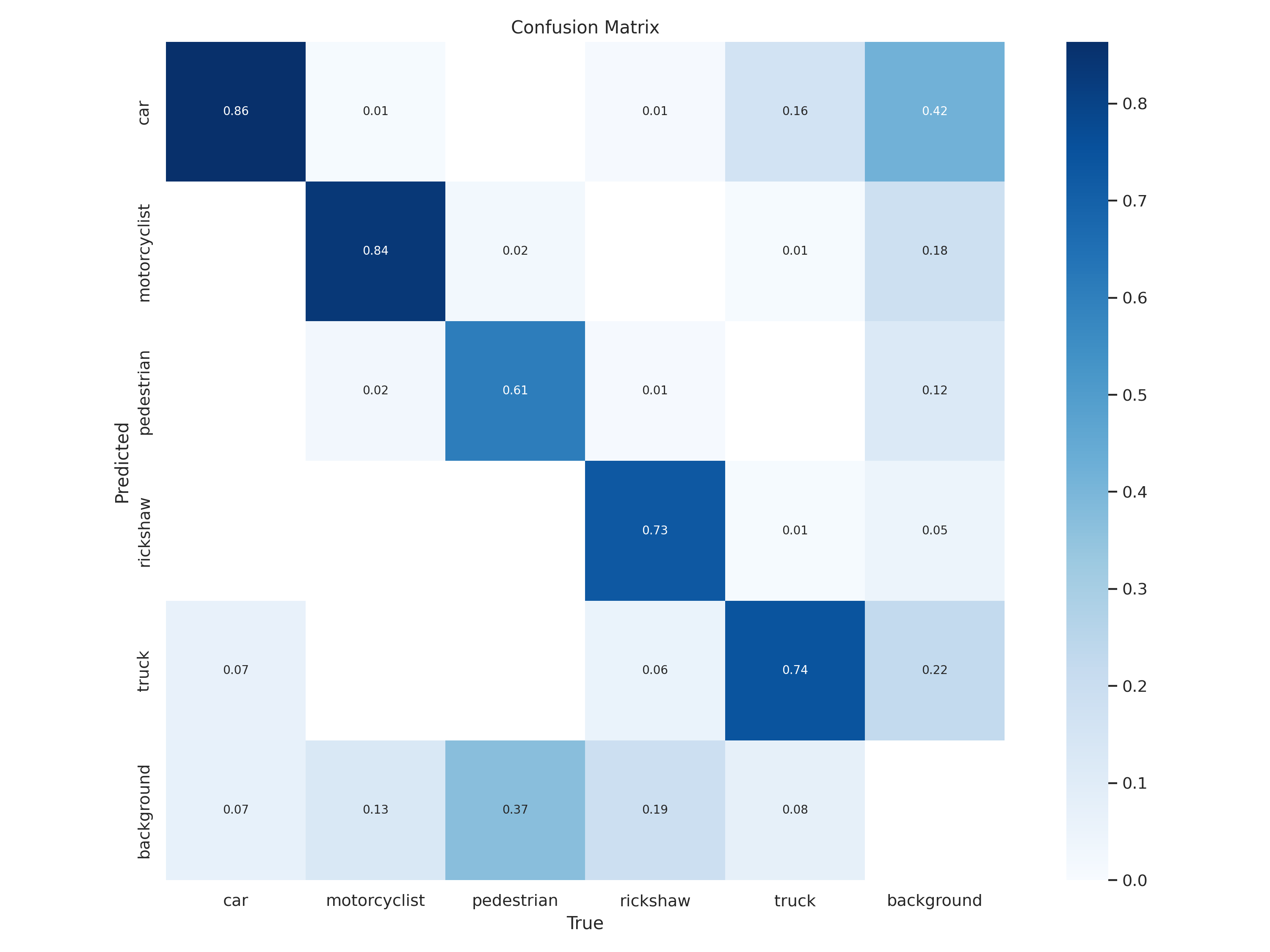}} \hfill
    \subfigure[Precision-Recall Curve.\label{fig:adam_e150_b16}]{\includegraphics[width=0.40\textwidth]{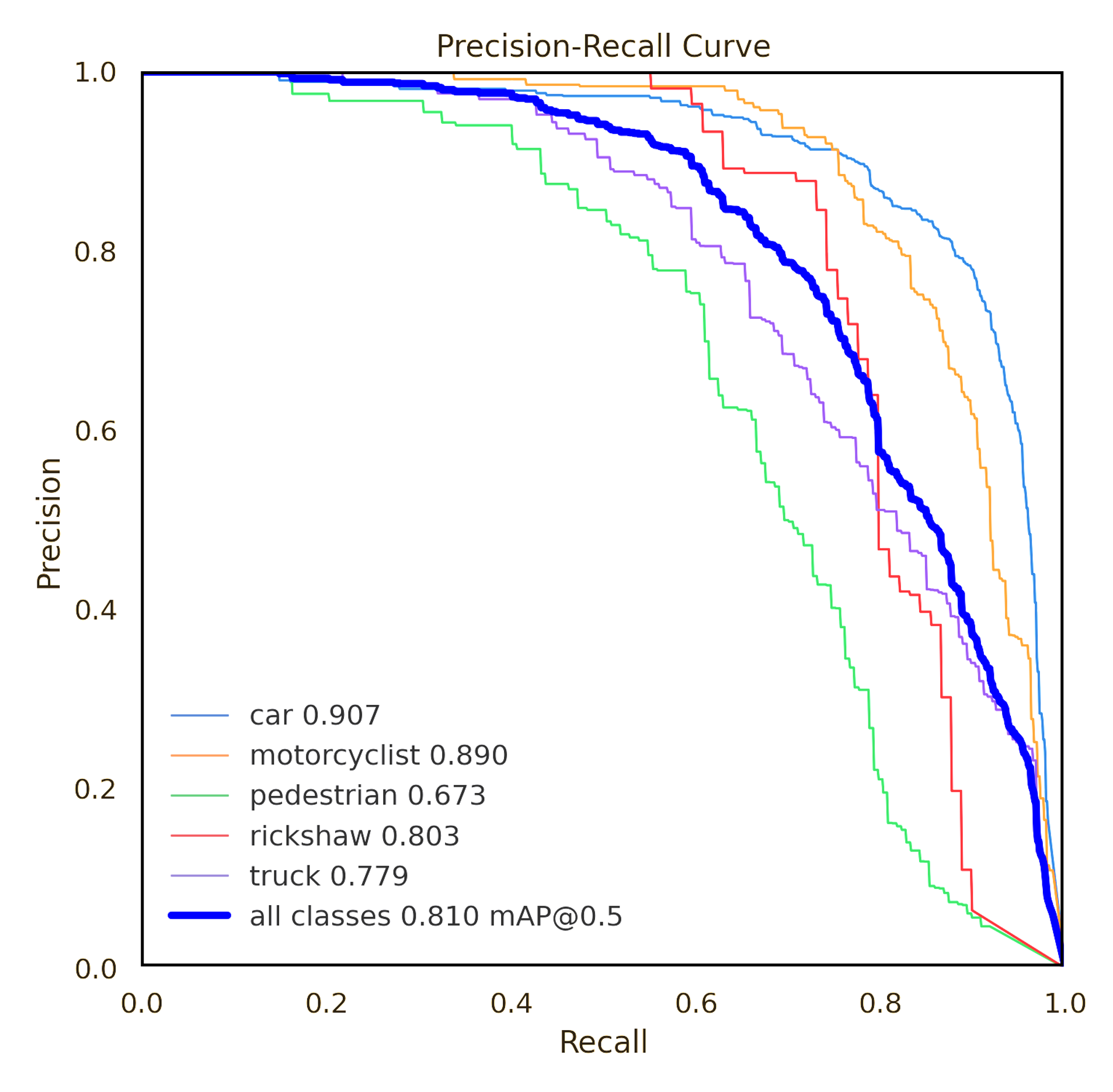}} \\[1ex]

    \caption{The Optimal YOLOv8s instance from Adam Group with 150 Epochs and 16 Batch-size.}
    \label{fig:adam_150_16}
\end{figure*}

\begin{figure*}[h]
    \centering
    \subfigure[Confusion Matrix.\label{fig:cm_rmsp_e100_b32}]{\includegraphics[width=0.50\textwidth]{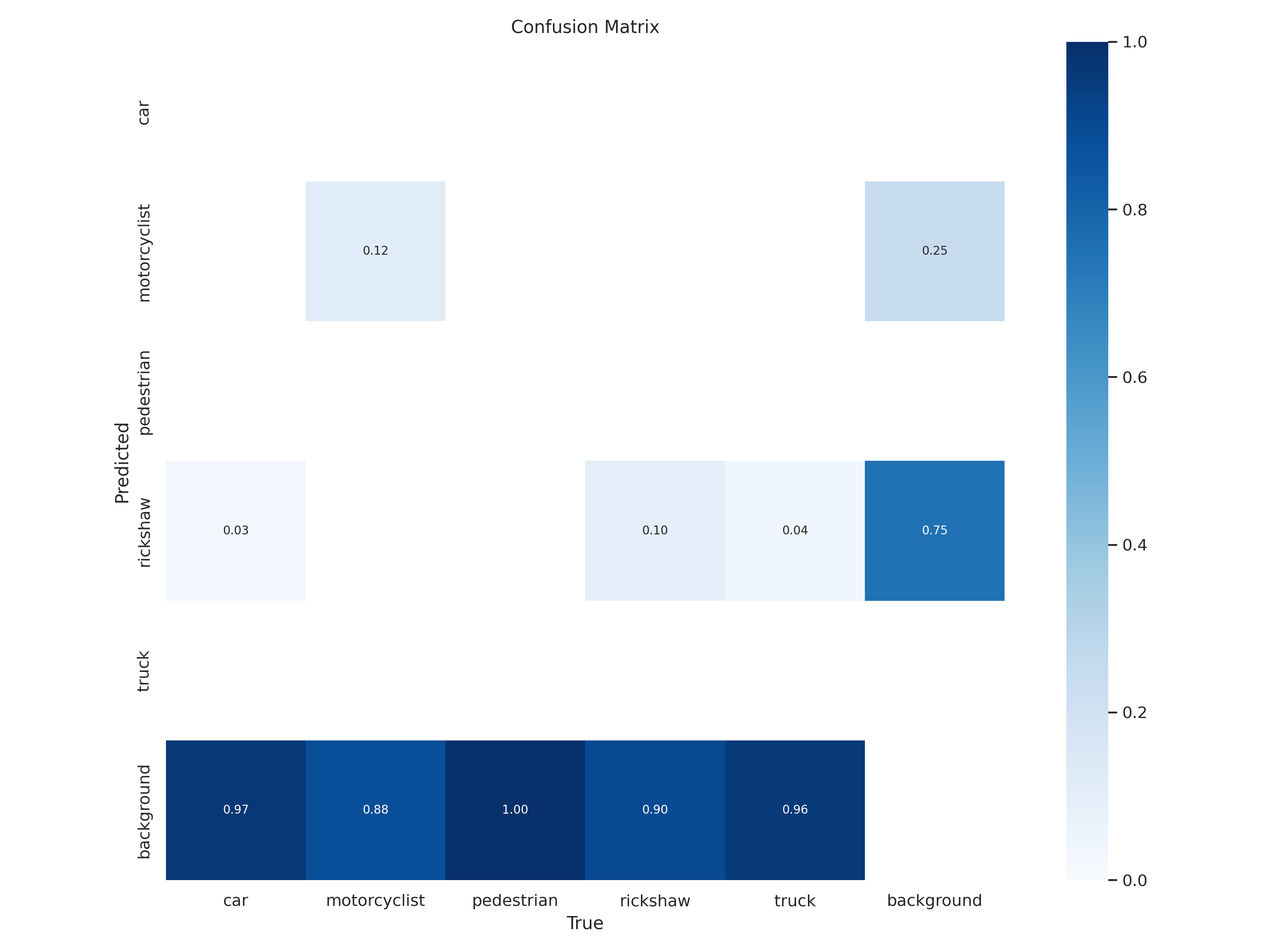}} \hfill
    \subfigure[Precision-Recall Curve.\label{fig:rmsp_e100_b32}]{\includegraphics[width=0.40\textwidth]{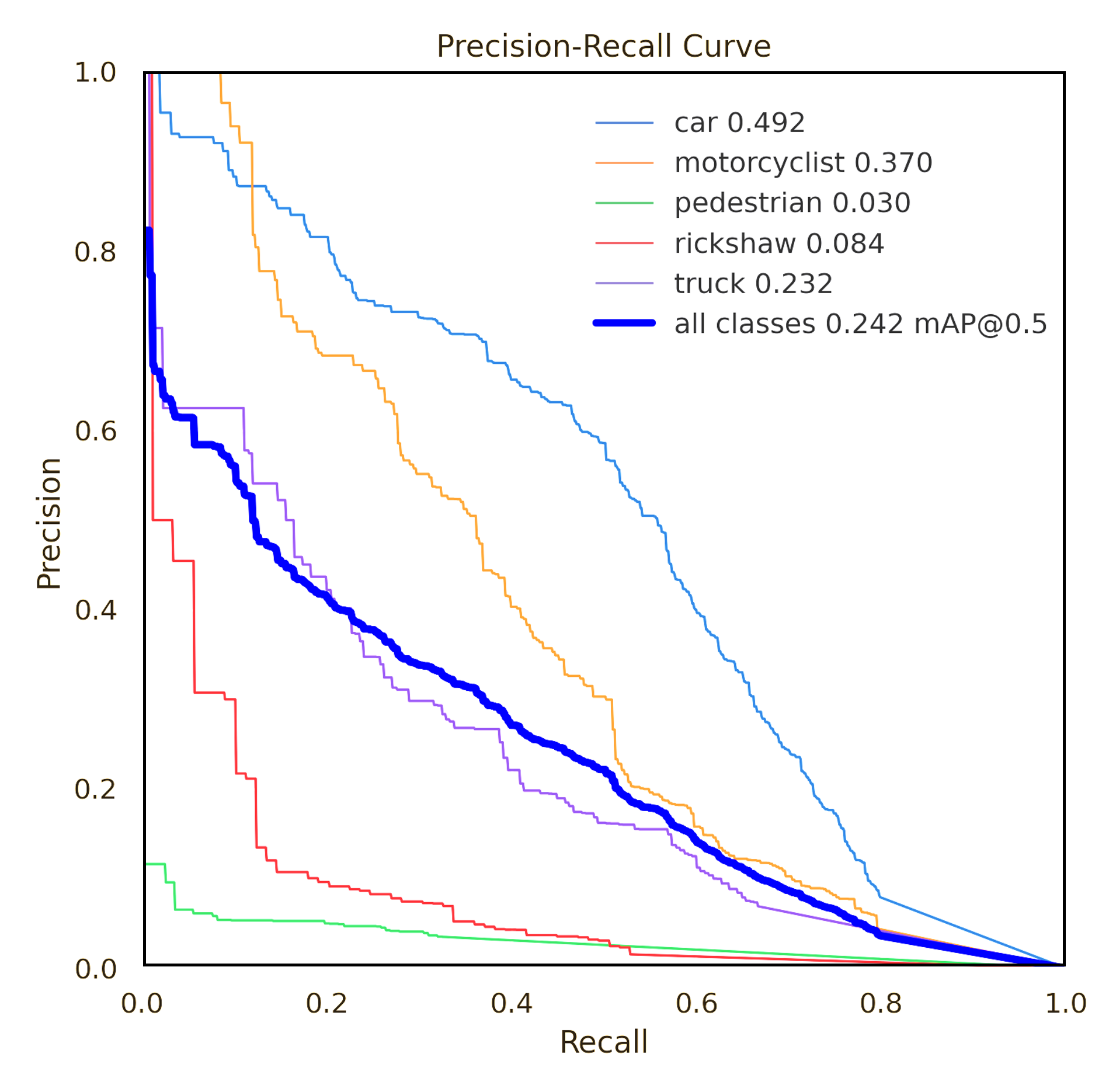}} \\[1ex]

    \caption{The Optimal YOLOv8s instance from RMSProp Group with 100 Epochs and 32 Batch-size.}
    \label{fig:rmsp_100_32}
\end{figure*}

\begin{figure*}[htbp]
    \centering
    \subfigure[Confusion Matrix.\label{fig:cm_adamw_e200_b32}]{\includegraphics[width=0.50\textwidth]{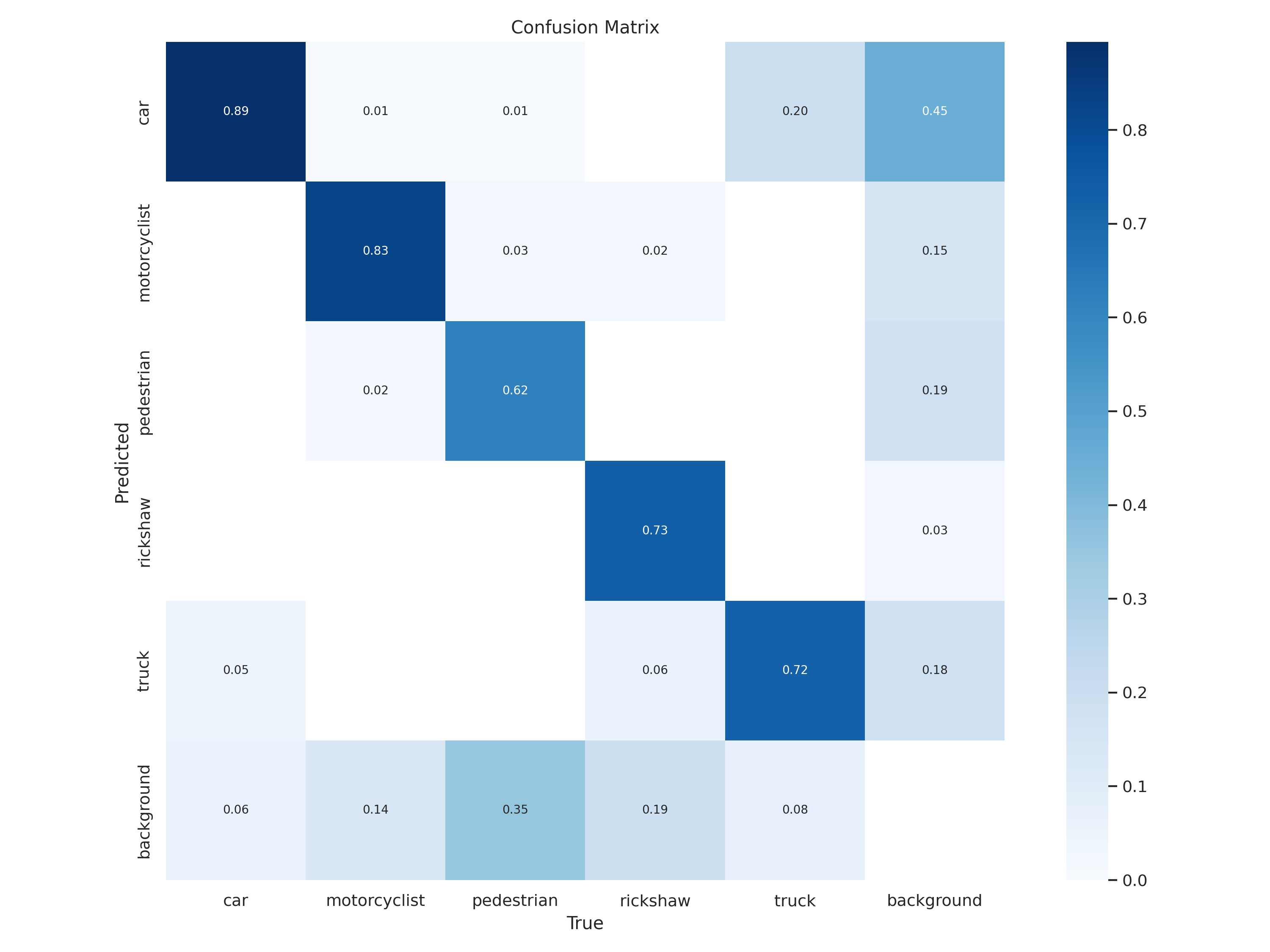}} \hfill
    \subfigure[Precision-Recall Curve.\label{fig:adamw_e200_b32}]{\includegraphics[width=0.40\textwidth]{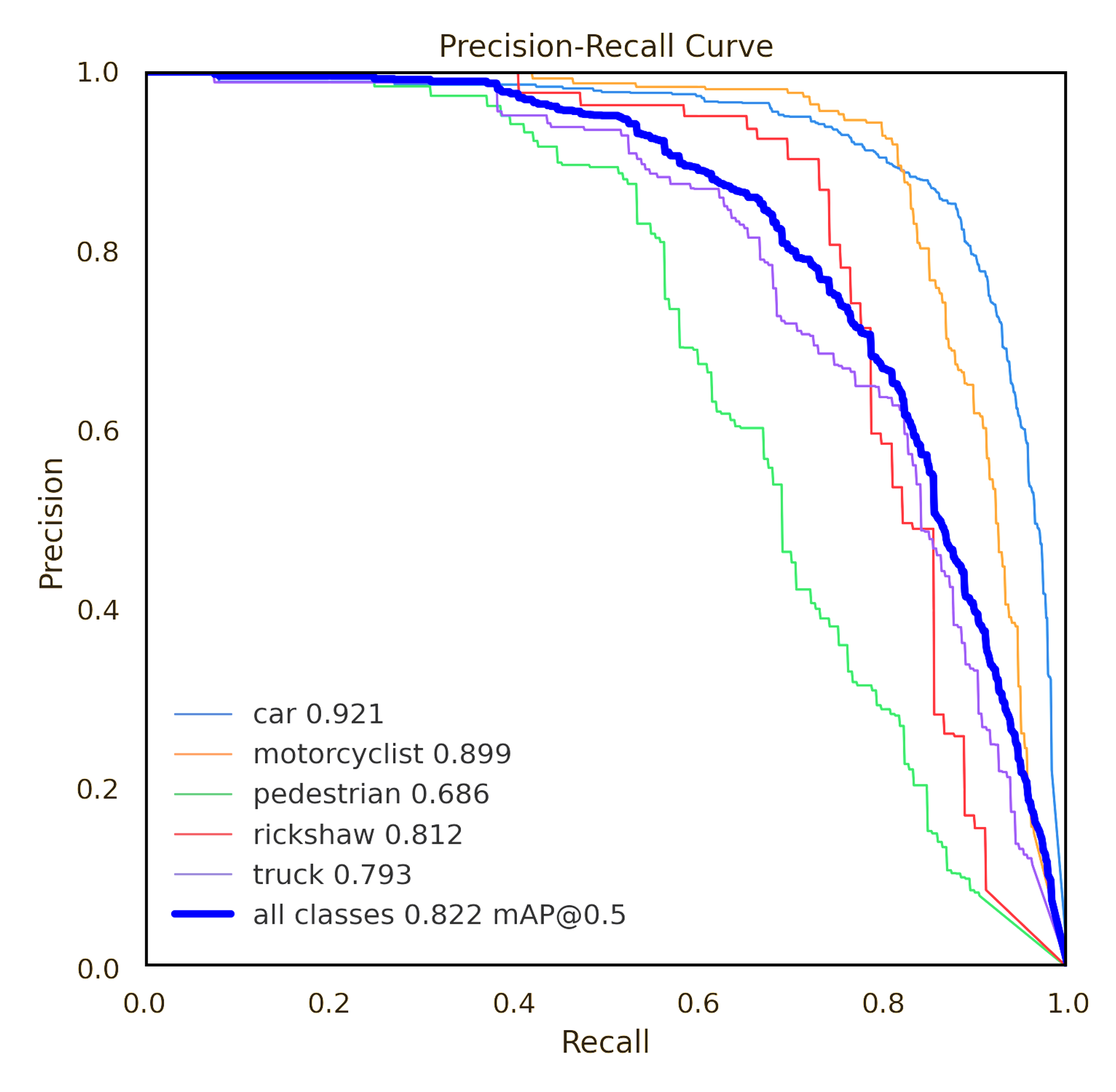}} \\[1ex]

    \caption{The Optimal YOLOv8s instance from AdamW Group with 200 Epochs and 32 Batch-size.}
    \label{fig:adamw_200_32}
\end{figure*}

The optimizer-based sets from Table~\ref{comparisonAllmAPs} can be used to consider four best-performing object detection models, i.e., YOLOv8s-3 from the SGD set, YOLOv8s-9 from the Adam set, YOLOv8s-14 from the RMSProp set, and YOLOv8s-24 from the AdamW set. More details about these models are provided in Table~\ref{comparisonAllInstances}, and the class-level details are provided in Table~A1, Table~A2, and Table~A3 in Appendix~\ref{app:details}. From a high-level perspective, the performance values reported for YOLOv8s-3 instance (best-performing object detection model instance from SGD group) in the confusion matrix, as shown in Figure~\ref{fig:cm_sgd_e150_b16}, achieved 89\% of car objects' detection, 84\% of \textit{motorcyclist} objects' detection, 67\% of \textit{pedestrian} objects' detection, 78\% of \textit{rickshaw} objects' detection, and 71\% of \textit{motorcyclist} objects' detection. A reflection of these results can also be witnessed in the PR-craves shown in Figure~\ref{fig:sdg_e150_b16}. These results show that the learning model instance may be suitable for any use-case scenario where the detection accuracy of at least 50\% is acceptable. Similarly, the YOLOv8s-9 instance from the Adam group can detect 86\% of car objects, 84\% of \textit{motorcyclist} objects, 61\% of \textit{pedestrian} objects, 73\% of \textit{rickshaw} objects, and 74\% of \textit{motorcyclist} objects. The confusion matrix and PR-curves can be seen in Figure~\ref{fig:adam_150_16}. Although the YOLOv8s-14 instance from the RMSProp group does not perform as well as its competing groups, it still performs well compared to other instances in its own group. The reason behind such behavior of RMSProp is as provided in Section~\ref{subsec:expResultsAndDiscussions}. We give the confusion matrix and PR-curves in Figure~\ref{fig:rmsp_100_32}. The last best performing YOLOv8s-24 instance is from the AdamW group, which can detect 89\% of \textit{car} objects, 83\% of \textit{motorcyclist} objects, 62\% of \textit{pedestrian} objects, 73\% of \textit{rickshaw} objects, and 72\% of \textit{motorcyclist} objects. The confusion matrix and PR-curves can be seen in Figure~\ref{fig:adamw_200_32}. 

Though the discussed four YOLOv8 model instances are best in their respective groups, we need to choose a single model instance among them. Since we know that a single mAP@0.5 value may not present the whole picture of performance, we rely on the well-known mathematical concept of "monotonic increase". The mAP@0.5 values of the AdamW group show a monotonic increase as compared to the mAP@0.5 values of the SGD group, as can be seen in Table~\ref{comparisonAllmAPs}. Therefore, we select \textit{YOLOv8s-24} as the optimal learning model instance for our detection-based perception system and relevant applications in AD. Now that we have fine-tuned the YOLOv8s model and have enough mAP performance values (as shown in Table~\ref{comparisonAllInstances}), we can substitute the performance values in Equation~\ref{odetFunc} and map the learning model instances' performances to perception satisfaction levels for autonomous vehicles.

\subsubsection{The Desired Perception Satisfaction Function}
\label{subsubsec:mAPs_to_psf}
In the previous section~\ref{subsec:fsfrs}, we have formulated the quantification of perception, performance metrics, threshold values, etc., and identified the necessary functions based on their behavior for our utility function. However, these attributes cannot define our desired perception satisfaction function unless the values of the underlying attributes are known. This can only be determined from the desired and considered real-world use case scenario for the perception system of the autonomous vehicle. For instance, the minimum and maximum threshold values originating from learning models' performance metrics will be different when several varying road segments, such as sharp turns, roundabouts, and straight roads, are considered for the perception system. Therefore, the choice of these attribute values is closely linked to the desired real-world use case scenario. 

Since we have considered detection-based perception satisfaction (as in Equation~\eqref{odetFunc}), we rely on the well-known public datasets for AD, such as KITTI~\cite{geiger2012we}, nuScenes~\cite{caesar2020nuscenes}, for their benchmarked learning models used for object detection and their performance metrics to be used as minimum and maximum thresholds for our desired perception satisfaction function. In this connection, we considered the top five listed learning models and their performance values from each dataset: VirConv-S (0.824)~\cite{VirConv, hailanyi82:online}, UDeerPEP (0.825)~\cite{TheKITTI32:online}, VirConv-T (0.812)~\cite{VirConv, hailanyi82:online}, HPC-Net (0.826)~\cite{TheKITTI70:online}, TSSTDet (0.806)~\cite{TheKITTI87:online} from KITTI's detection task~\cite{TheKITTI52:online}; Far3D (0.635), HoP (0.624)~\cite{zong2023temporal}, Li (0.623), StreamPETR (0.620)~\cite{wang2023exploring}, SparseBEV (0.603)~\cite{liu2023sparsebev} from nuScenes's detection task~\cite{Objectde0:online}. The source codes can be found for: HoP~\cite{SenseXHo37:online}, StreamPETR~\cite{exiawshS14:online}, and SparseBEV~\cite{MCGNJUSp4:online}. Hence, we choose our minimum and maximum threshold values to be equal to \textit{0.6} and \textit{0.8}, respectively. Now we can draw our transition region and adjust its length using the $beta$ value to aid the convergence of perception satisfaction into a delighted state. It is worth noting that our selection of threshold values is based on the performance of the above-listed SOTA learning models used for real-time object detection, and the choice of these threshold values is closely tied to the underlying real-time use case scenarios. Now that we have the attribute values and can draw our \textit{desired} perception satisfaction function, we can utilize our trained YOLOv8s model instances and their performance values. 

\begin{figure*}[]
\centering
\includegraphics[width=\textwidth]{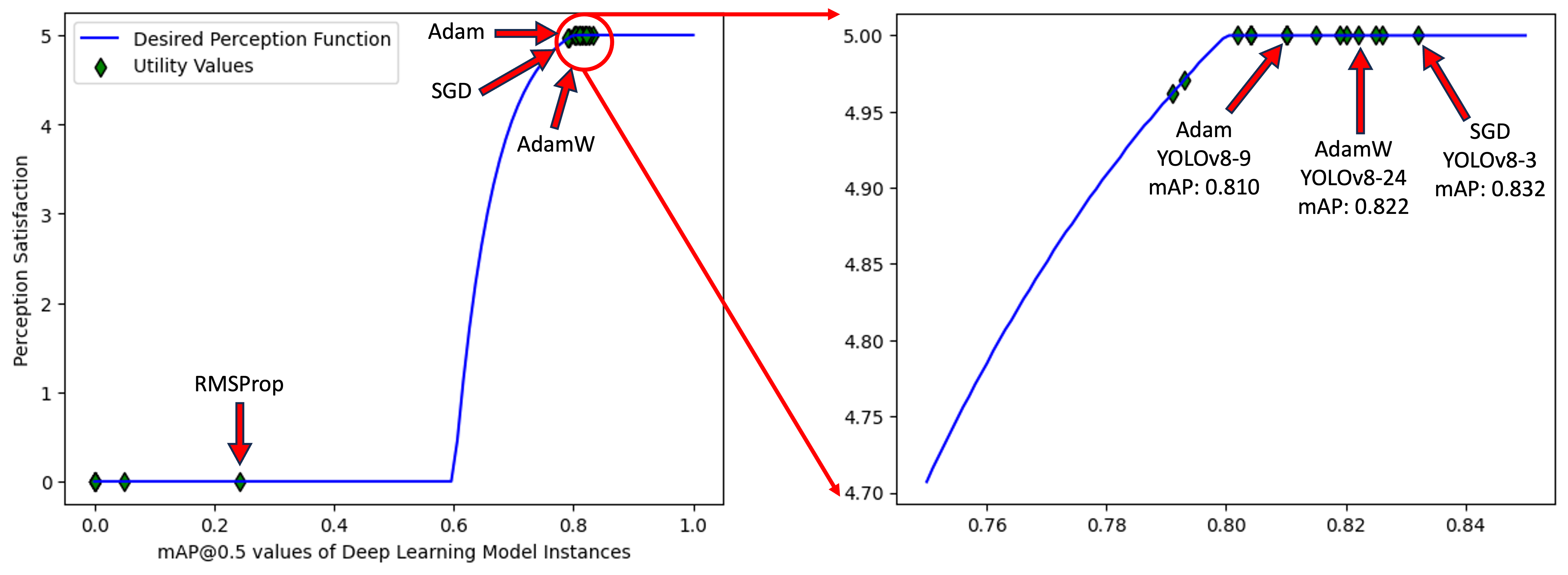}
\caption{The mAP@0.5 values.}
\label{mAP_50}
\end{figure*}

\begin{figure*}[]
\centering
\includegraphics[width=\textwidth]{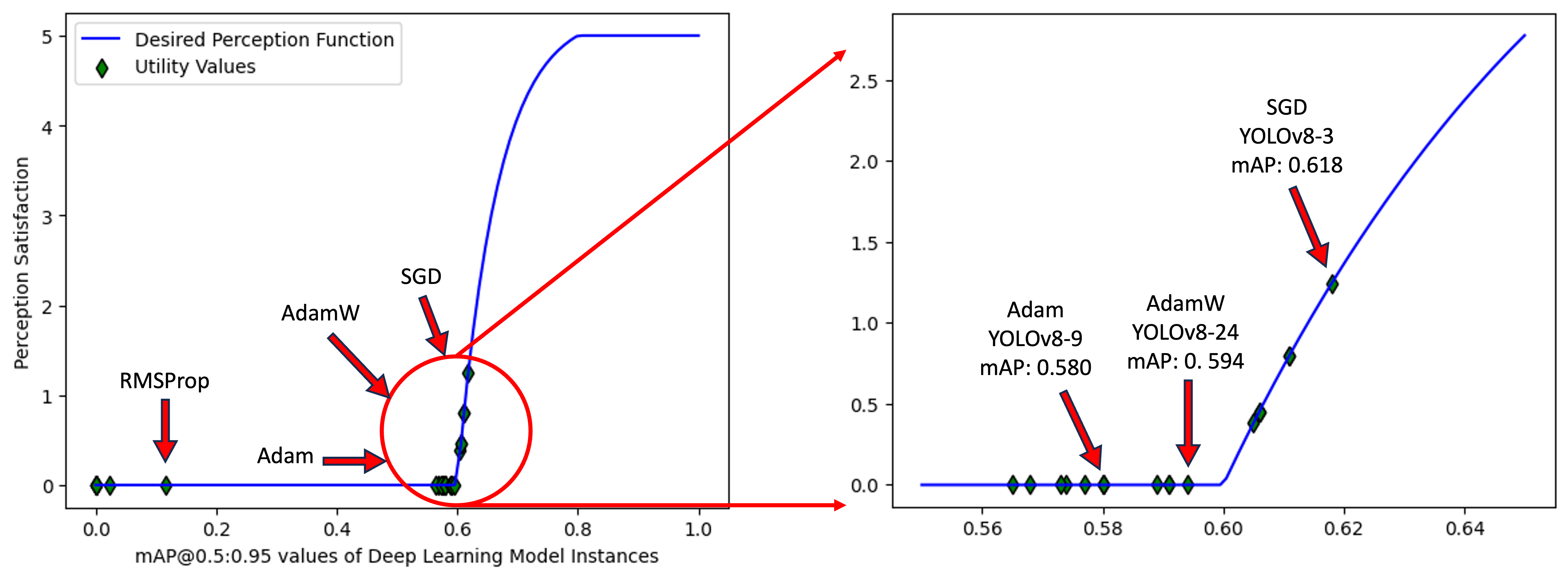}
\caption{The mAP@0.5:0.95 values.}
\label{mAP_50_95}
\end{figure*}

The utility values for our YOLOv8s model instances based on mAP@0.5 performance values can be seen in Figure~\ref{mAP_50}. The desired perception function curve shows that almost all instances perform well except the RMSProp set of instances. The reasons for this behavior of RMSProp instances are documented in Section~\ref{subsec:expResultsAndDiscussions}. A detection-based perception system using our satisfaction function would never recommend such learning models for object detection in autonomous vehicles. A close investigation of the high-performing model instances in the delighted state confirms that YOLOv8s-9, YOLOv8s-24, and YOLOv8s-3 are the best-performing instances from the Adam, AdamW, and SGD sets, respectively. Though it affirms that AdamW outperforms Adam, it contradicts our claim at the current stage (mAP@0.5) that AdamW is an optimal instance due to monotonic increasing behavior. Overall, it confirms that these three learning model instances qualify for selection in the perception system for detection tasks in any non-critical applications in AD. For critical applications, we rely on well-known mAP@0.5:0.95 performance values of our YOLOv8s model instances. The satisfaction utility for these instances is shown in Figure~\ref{mAP_50_95}. A close investigation of the acceptable-performing model instances between the entirely unsatisfied and fully satisfied regions confirms that even the best-performing models may not be acceptable for perception satisfaction when the underlying real-world use case scenario and nature of applications change. In addition, the SGD (YOLOv8s-3) model instance is performing well, which again contradicts our claim about AdamW (YOLOv8s-24) at the current stage (mAP@0.5:0.95).

\begin{figure*}[]
\centering
\includegraphics[width=\textwidth]{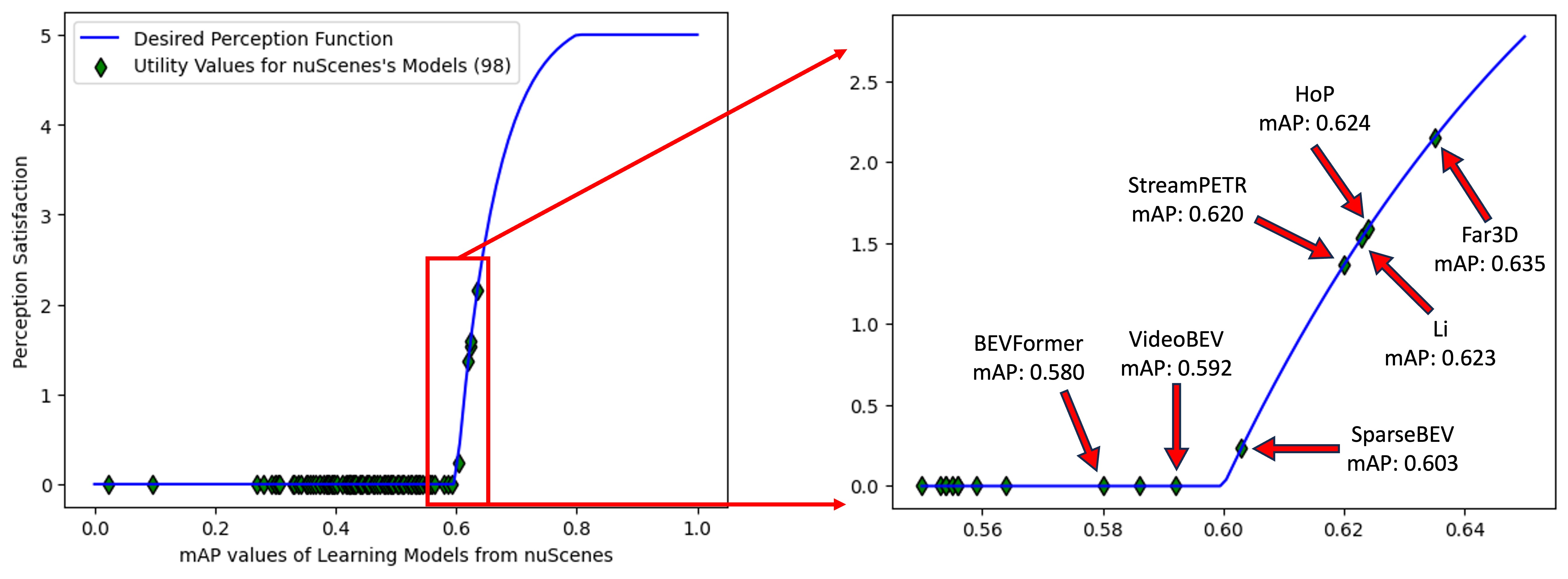}
\caption{The mAP values from nuScenes.}
\label{mAP_nuScenes}
\end{figure*}

To confirm the optimal performance of our proposed utility function, we rely on the well-known concept of \textit{benchmarking} to evaluate the process of finding utility values for well-known detection-based learning models. Therefore, we selected all real-time object detection-based learning models benchmarked for large public nuScenes dataset and plotted their satisfaction utility values over our desired perception satisfaction function as shown in Figure~\ref{mAP_nuScenes}. A close investigation of the curve indicates that only a few learning models are in the region of acceptable satisfaction. For instance, Far3D (0.635), HoP (0.624)~\cite{zong2023temporal}, Li (0.623), StreamPETR (0.620)~\cite{wang2023exploring}, and SparseBEV (0.603)~\cite{liu2023sparsebev} can be considered by the perception system if the underlying use case scenario requires at least 60\% of the object detection accuracy. Therefore, the perception system of an autonomous vehicle may reject even an emerging transformation model i.e., BEVFormer (0.580)~\cite{yang2023bevformer}. It is worth noting that mAP values used by detection models of nuScenes dataset are different than those determined by us in the form of mAP@0.5 and mAP@0.5:0.95. Therefore, to mitigate the risk of bias, we rely on the tight performance values (as shown in the extended part of Figure~\ref{mAP_50_95}) for comparing the process of finding utility values. Meaning thereby, we can rely on our trained YOLOv8s-3 model instance when the real-world use case scenario matches with those of our learning model and the nuScenes models. 

\begin{figure}[b]
\centering
\includegraphics[width=8cm]{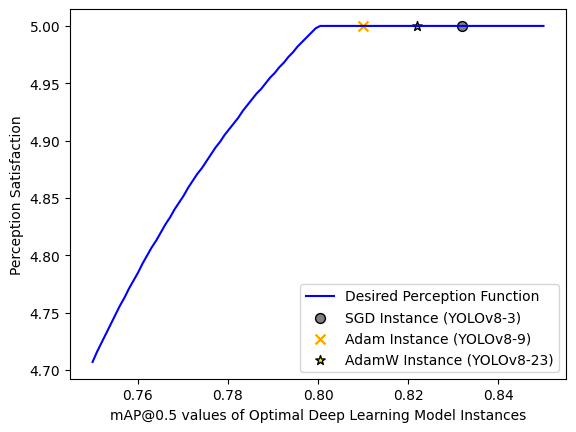}
\caption{Optimal Model Instances for thresholds [0.6 - 0.8].}
\label{optimal_model_instances}
\end{figure}

\begin{figure}[]
\centering
\includegraphics[width=8cm]{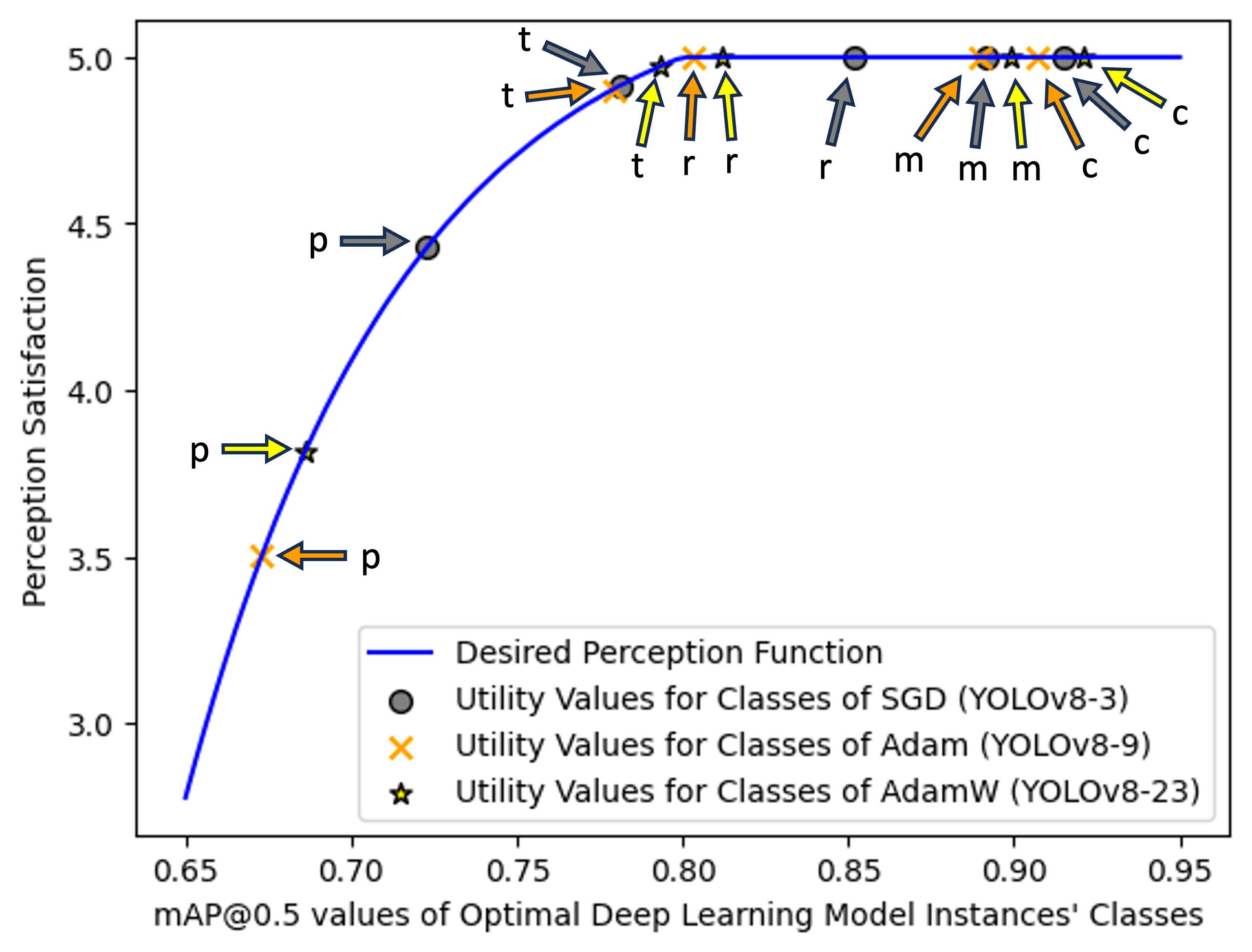}
\caption{Optimal Model Classes for thresholds [0.6 - 0.8].}
\label{optimal_model_classes}
\end{figure}

Now that we have benchmarked the process of finding utility values, we worked on mitigating the contradiction caused by SGD (YOLOv8s-3) and AdamW (YOLOv8s-23). The YOLO8-3 model instance outperforms YOLOv8s-24 model instance as shown in Figure~\ref{optimal_model_instances}. However, we know that a single mAP@0.5 performance value cannot truly represent the actual performance of a learning model. A detailed discussion and relevant reasons are provided in Section~\ref{subsubsec:yolov8_1_model}. As discussed earlier, we relied on granular analysis and performed a deep examination of the individual class performance of these learning models. A close investigation of the curve shows that the majority of class-level performance values confirm that YOLOv8s-24 model instance outperforms the YOLOv-3 model instance, as shown in Figure~\ref{optimal_model_classes}. For instance, the car (c) class, motorcyclist (m) class, and truck (t) class outperform the class-level performance of both YOLOv8s-3 and YOLOv8s-9 model instances. In addition, the rickshaw (r) class and pedestrian (p) class performance of YOLOv-24 outperforms the class-level performance of the YOLOv8s-9 model instance. It is worth noting that these three classes have a total of \textit{6002} number of objects, which represent 81.46\% of objects in our custom dataset. Therefore, the contradiction is removed based on our class-level findings.  

\subsubsection{The Worldview of Perception Satisfaction Function}
\label{subsubsec:world_view_psf}

\begin{figure}[]
\centering
\includegraphics[width=8cm]{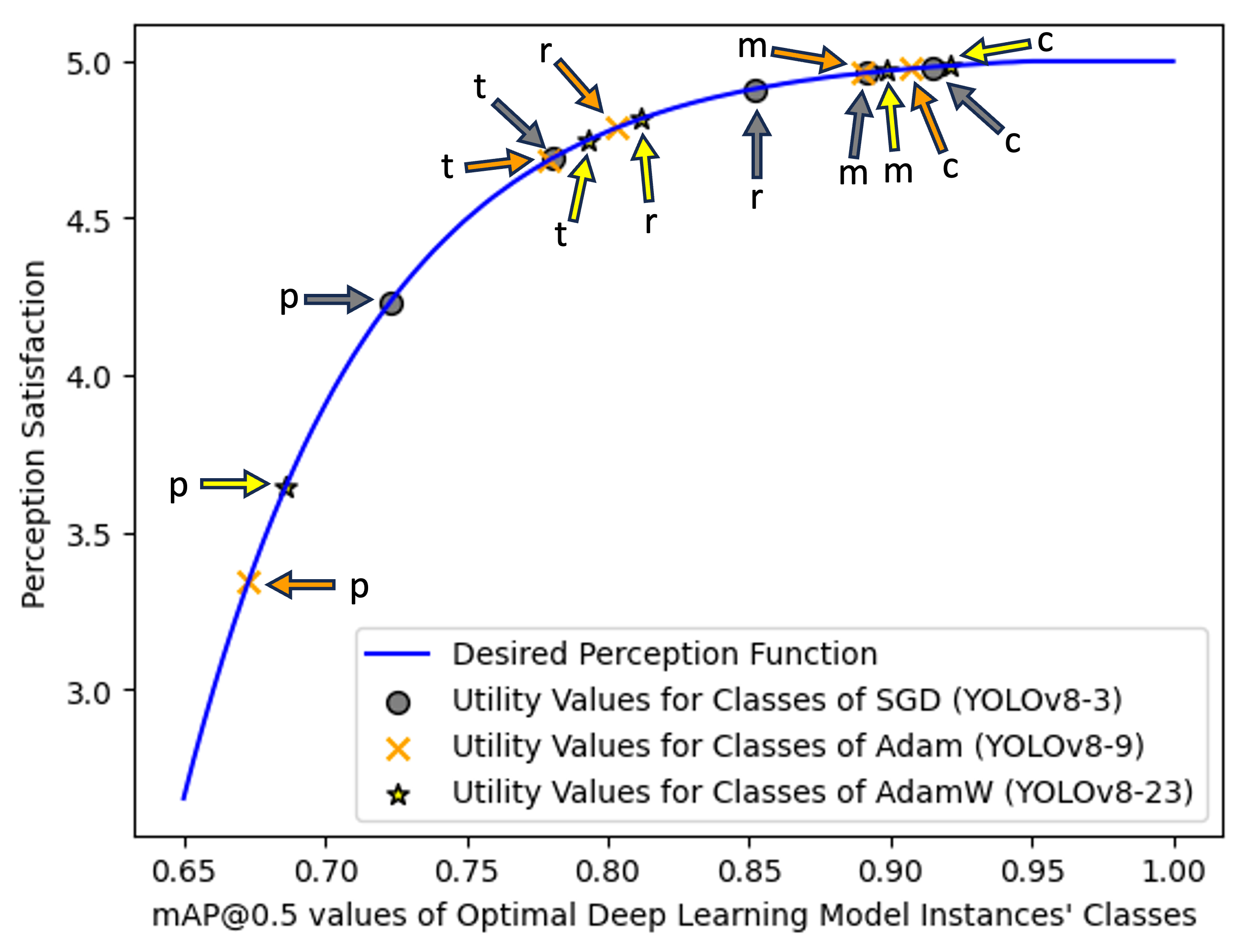}
\caption{Optimal Model Classes for thresholds [0.6 - 0.95].}
\label{general_optimal_model_classes}
\end{figure}

Our proposed analytical function focused on how autonomous vehicles assess the utility of their considered machine learning and DL models. It is worth noting that all attribute values used in our satisfaction function are tailored to our use in this paper (based on the SOTA performance values of the well-known models). Therefore, the research community and relevant practitioners may utilize different attribute values considering a broader worldview. Any SOTA machine learning and DL model can plug its performance values, along with any combination of threshold values, into our analytical function to find the utility values. A close investigation of the desired curve (where the maximum threshold changed from \textit{0.8} to \textit{0.95}) shows similar results, i.e., YOLOv8s-24 outperforms the class-level performance as shown in Figure~\ref{general_optimal_model_classes}. However, the effects of changing the maximum threshold can be seen by all classes lying in the acceptable satisfaction area rather than the fully satisfied state for AV. Therefore, our worldview of the perception satisfaction function accepts the performance values of any learning model and provides the utility for the optimal perception selection for the autonomous vehicle. Nevertheless, our findings also suggest that learning models can be chosen based on the specific satisfaction of class-level detection, aligning with underlying real-world use case scenarios for AD.

\section{Conclusion}
\label{sec:concl}
Autonomous vehicles are faced with the challenge of selecting the correct perception produced by increasingly growing DL models. We proposed a novel utility-based analytical perception satisfaction function to evaluate the performance of learning models used for accurate perception creation in AVs. For this, we acquire a custom dataset with multiple objects, such as a car, pedestrian, motorcyclist, truck, and rickshaw. For the detection-based perception task, we rely on a well-known SOTA DL-based YOLOv8s model. We experimented with the YOLOv8s model and trained 24 different instances with our custom dataset. Furthermore, we used the performance values of these learning model instances and top SOTA benchmarked models from two well-known public datasets to obtain the utility values for vehicle-level satisfaction. Our perception function validation and experimental process demonstrate how the SGD-based YOLOv8s model instance was erroneously considered suitable for accurate and correct perception, based on its performance value (mAP@0.5: 0.832). This misconception has been clarified through the lens of our perception satisfaction function, which utilizes class-level performance and finds that the AdamW-based YOLOv8s model instance is most suitable for accurately perceiving cars, motorcyclists, and trucks. The research community and practitioners can find our satisfaction function and research results useful when using available algorithms and models to determine the appropriate perception for AVs. The applications of our proposed satisfaction function extend beyond the documented perception task(s), as they can be applied to evaluate any perception service in AVs. In future work, we plan to assess the proposed utility function through human-in-the-loop validation and benchmark it using different SOTA object detection models.

\begin{appendices}
\section{\break Confusion Matrices and PR-Curves}
\label{app:details}
This appendix provides individual class performances, confusion matrices, and PR curves of the YOLOv8s model instances. We are providing the corresponding tables (A1-A3) and figures (A1-A6) online through \href{https://drive.google.com/drive/folders/1YOYHWf4H3Pem3jjO48tmbFIou-nKQg4g?usp=drive_link}{\underline{Google Drive}}.
\end{appendices}

\section*{Acknowledgment}
This work was supported in part by the Emirates Center of Mobility Research (ECMR) UAEU, in part by Sandooq Al Watan (SW) UAE, in part by ASPIRE Award for Research Excellence (AARE20-368), and in part by UAE University and Zayed University (UAEU-ZU) research project, UAE.

\end{document}